\documentclass[lettersize,journal]{IEEEtran}
\usepackage{amsmath,amsfonts}
\usepackage{algorithmic}
\usepackage{algorithm}
\usepackage{array}
\usepackage{textcomp}
\usepackage{stfloats}
\usepackage{url}
\usepackage{verbatim}
\usepackage{graphicx}
\usepackage{amssymb}
\usepackage{mathtools}
\usepackage{amsthm}
\usepackage{bm}
\usepackage[american]{babel}
\usepackage{hyperref}
\mathtoolsset{showonlyrefs,showmanualtags}
\usepackage{color}

\usepackage{xcolor}  
\newcommand{\fix}[1]{#1}
\usepackage{amsmath}
\usepackage{amssymb}
\usepackage{times}
\usepackage{soul}
\usepackage{url}
\usepackage[utf8]{inputenc}
\usepackage{amsthm}
\usepackage{booktabs}
\usepackage{algorithmic}
\usepackage[switch]{lineno}
\usepackage{multirow}
\usepackage{bm}
\usepackage{subfig}

\begin{document}

\clearpage
\newpage

\title{Heterogeneous Multi-agent Zero-Shot Coordination by Coevolution}




\author{Ke Xue, Yutong Wang, Cong Guan, Lei Yuan, Haobo Fu, Qiang Fu, \\ Chao Qian,~\IEEEmembership{Senior Member,~IEEE,} Yang Yu,~\IEEEmembership{Senior Member,~IEEE} 
\thanks{
This work was supported by the National Science and Technology Major Project (2022ZD0116600), National Science Foundation of China (62276124), Jiangsu Science Foundation (BK20243039, BK2024119), and the Fundamental Research Funds for the Central Universities (14380020). Chao Qian is the corresponding author (email: qianc@nju.edu.cn).

Ke Xue, Yutong Wang, Cong Guan, Lei Yuan, Chao Qian and Yang Yu are with the National Key Laboratory for Novel Software Technology, and also with School of Artificial Intelligence, Nanjing University, Nanjing 210023, China.

Haobo Fu and Qiang Fu are with the Tencent AI Lab, Shenzhen 518000, China.
}
}

\markboth{IEEE Transactions on Evolutionary Computation,~Vol.~xx, No.~x, ~2023}%
{Shell \MakeLowercase{\textit{et al.}}: A Sample Article Using IEEEtran.cls for IEEE Journals}
\maketitle

\begin{abstract}
Generating agents that can achieve zero-shot coordination (ZSC) with unseen partners is a new challenge in cooperative multi-agent reinforcement learning (MARL). Recently, some studies have made progress in ZSC by exposing the agents to diverse partners during the training process. They usually involve self-play when training the partners, implicitly assuming that the tasks are homogeneous. However, many real-world tasks are heterogeneous, and hence previous methods may be inefficient. In this paper, we study the heterogeneous ZSC problem for the first time and propose a general method based on coevolution, which coevolves two populations of agents and partners through three sub-processes: pairing, updating and selection. Experimental results on various heterogeneous tasks highlight the necessity of considering the heterogeneous setting and demonstrate that our proposed method is a promising solution for heterogeneous ZSC tasks. \fix{To the best of our knowledge, we are the first to underscore the significance of the heterogeneous ZSC tasks and to introduce an effective framework for addressing it.} 
\end{abstract}

\section{Introduction} \label{introduction}

\IEEEPARstart{R}einforcement learning (RL)~\cite{sutton2018reinforcement} has gained many successes against humans in competitive games, such as Go~\cite{silver2016mastering},  StarCraft~\cite{vinyals2017starcraft} and Dota~\cite{berner2019dota}. However, developing Artificial Intelligence (AI) that can coordinate with human partners (i.e., Human-AI coordination) is emerging as a new challenge in cooperative multi-agent RL (MARL)~\cite{yuan2023survey}. Achieving zero-shot coordination (ZSC)~\cite{ZSC,off-belief,fcp} is especially important when an agent must generalize to novel human partners that are unseen during its training phase.

In the problem of ZSC, a trained agent is required to coordinate with novel partners (i.e., partners unseen during training)~\cite{ZSC,adhoc-survey}. To achieve this goal, previous methods mainly expose the agent to diverse partners during the training process. Specifically, they usually obtain a set of diverse partners by individual-based or population-based self-play~\cite{sp,sp2}, i.e., a policy acts as both an agent and a human partner simultaneously, then use the set of diverse partners to train a robust agent~\cite{TrajeDi,fcp,MEP}. Note that the validity of obtaining diverse partners via self-play relies on the assumption that the agents and partners are homogeneous, i.e., a policy can be treated as both an agent and a partner.

\begin{figure}[t]
    \centering
    \includegraphics[width=0.49\textwidth]{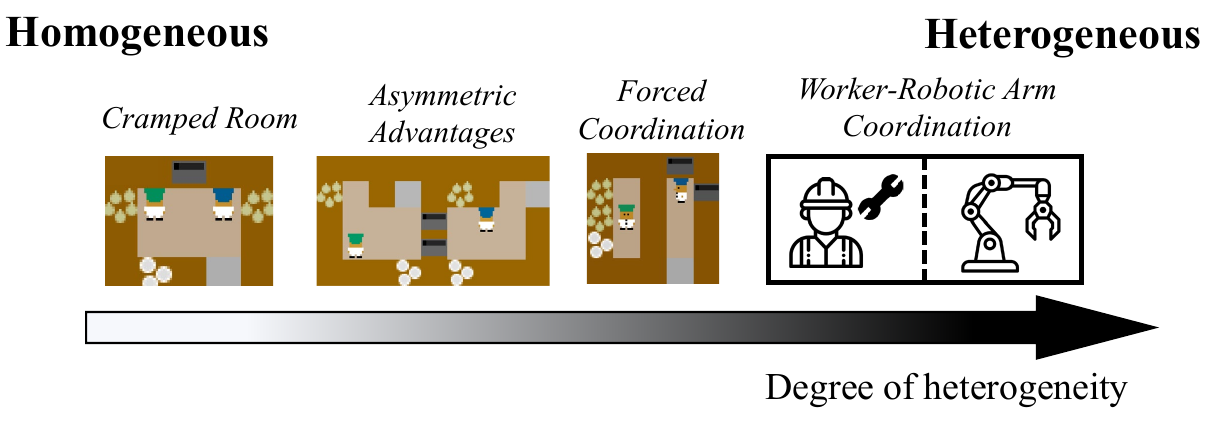}
    \caption{An example illustration of Human-AI coordination with increasing heterogeneity.}
    \label{the degree of heterogeneous}
\end{figure}

In many real-world tasks, AI and humans are, however, usually heterogeneous~\cite{heterogenous-early,robot-arm-lifeifei}, as shown in Figure~\ref{the degree of heterogeneous}. That is, agents and partners have different advantages (i.e., the efficiency of doing something) or skills (i.e., the ability to do something). \fix{For example, in factory production tasks, AI agents (e.g., robotic arms) can lift heavy objects, while human workers can perform more flexible operations than the robotic arm. In disaster search and rescue missions, AI agents (e.g., drones and quadruped robots) exhibit faster mobility and are better suited to explore dangerous, hard-to-reach areas, while human doctors have slower speed but can better treat the injured.}
The heterogeneous setting has greater applicability and can recover homogeneous policies as a result if necessary~\cite{heterogeneous-mirror}. Though heterogeneous tasks are widely available and important, it is nontrivial to train heterogeneous agents, especially in the ZSC setting. \fix{Previous self-play methods mainly focus on homogeneous settings, which may not capture the cooperative behaviors between agents and human partners well, thus being inefficient for heterogeneous ZSC tasks. There is a significant gap between the demands of real-world applications and the deficiencies in heterogeneous algorithm research.} Thus, a natural question is raised: \emph{Can we generate agents without human data that can coordinate with novel and heterogeneous humans efficiently?}

In this paper, we propose a general method, Multi-Agent Zero-shot coordination by coEvolution (MAZE), for the heterogeneous ZSC task. Instead of training agents and partners by self-play, MAZE explicitly employs two populations to maintain agents and partners, respectively, and coevolves the two populations throughout the training phase. The effectiveness of considering heterogeneous agents separately has been verified by recent theoretical analysis in heterogeneous MARL~\cite{heterogeneous-jmlr,hu2023heterogeneous}. To enable the agent can coordinate well in the ZSC setting, MAZE exposes the agent to diverse partners~\cite{TrajeDi,fcp,MEP}, which is achieved by pairing, updating, and selection. In each generation, the agents and partners from the two populations are first paired to coordinate in the environment and then updated by optimizing a weighted sum of reward and diversity. To further enhance the diversity of partners, MAZE maintains an archive storing diverse partners generated so far, from which some past partners are selected.

We mainly evaluate the performance of MAZE in the popular \emph{Overcooked}~\cite{overcooked} environment, a two-player common-payoff collaborative cooking environment. Furthermore, we design a grid-world \emph{FillInTheGrid} to verify the versatility of MAZE. We conduct experiments on different layouts in these environments, where the agent and partner have different degrees of heterogeneity. We investigate four research questions (RQs) in our experiments. In RQ1, we show that even a simplified variant of MAZE, which just uses two populations of agents and partners, can be significantly better than self-play~\cite{sp,sp2} and population play~\cite{pp}, disclosing the necessity of considering the heterogeneity of agent and partner. The results in RQ2 show that MAZE achieves better performance compared with recently proposed methods~\cite{TrajeDi,fcp,MEP}. In RQ3, we examine the influence of different components of MAZE, demonstrating the effectiveness of each component of the proposed framework. In RQ4, we investigate the coordination ability with real human participants. To the best of our knowledge, we are the first to point out the importance of heterogeneity in ZSC and propose an efficient method MAZE to solve it. Experiments on different types of heterogeneous environments show the necessity of considering the heterogeneity and the effectiveness of MAZE.

\fix{Our contributions are three-folds:

\begin{itemize}
   \item \textbf{Introduction of a Novel Problem Setting}: We pioneer the exploration of heterogeneous ZSC. This paper is the first to systematically identify the complexities and necessities of addressing heterogeneity in ZSC, demonstrating its prevalence and significance in many real-world scenarios.
   \item \textbf{Development of an Effective Framework}: We introduce MAZE, a novel framework specifically designed for heterogeneous ZSC challenges. Unlike traditional approaches that rely on homogeneous self-play, MAZE innovatively employs a coevolutionary strategy involving two distinct populations of agents and partners. This method not only enhances the adaptability and robustness of AI agents when coordinating with novel, diverse human partners but also effectively addresses the inherent challenges posed by the heterogeneity of capabilities and roles. 
   \item \textbf{Comprehensive Empirical Validation and Analysis}: Through rigorous experimentation in the Overcooked environment and a custom-designed grid-world setting, FillInTheGrid, we provide a thorough empirical evaluation of MAZE. Our analysis demonstrates the superiority of MAZE over existing methods in the heterogeneous ZSC tasks. 
\end{itemize}
}

\section{Background}
A general formulation for fully cooperative MARL is Decentralized Partially Observable Markov Decision Process (Dec-POMDP)~\cite{dec-pomdp,xue2022mis}, which is denoted as a tuple $\langle \mathcal{I}, \mathcal{S}, \bm{\mathcal{A}}, \mathcal{P}, \bm{\Omega}, \bm{O}, R, \gamma\rangle$, 
where $\mathcal{I}=\{1,\dots,N\}$ indicates the set of $N$ controlled agents, $\mathcal{S}$ is the state space, $\bm{\mathcal{A}} = \mathcal{A}^{1}\times \cdots \times \mathcal{A}^{N}$ is the joint action space, $\mathcal{P}: \mathcal{S} \times \boldsymbol{\mathcal{A}} \to \mathcal{S}$ is the transition function, 
$\bm{\Omega}$ is the joint observation space, $\bm{O}$ is the joint observation function space,
$R: \mathcal{S}\times\boldsymbol{\mathcal{A}}\to\mathbb{R}$ is a shared global reward function, and $\gamma\in [0,1)$ is the discount factor. 
At each time-step, each agent $i$ observes a local observation $o_i\in \Omega^i$, which is a projection of the true state $s\in \mathcal{S}$ by the observation function $o_i=O^i\left(s,i\right)$. Each agent selects an action $a_i\in \mathcal{A}$ to execute, and all individual actions form a joint action $\bm{a}\in\mathcal{A}^N$ which leads to the next state $s'\sim \mathcal{P}\left(s'|s,\bm{a}\right)$ and a shared reward $r=R\left(s,\bm{a}\right)$, the formal objective of the agents is to maximize the expected cumulative discounted reward $\mathbb{E}[\sum_{t=0}^\infty \gamma^t R(s_t, \pmb{a}_t)]$.

Zero-shot coordination (ZSC) aims to train agents (representing AI) that can coordinate well with novel partners (representing humans)~\cite{ZSC,off-belief}, which can be seen as a special type of ad hoc teamwork problems~\cite{adhoc-aaai10,adhoc-survey}. It can be modeled as a Dec-POMDP, denoted by $\langle \mathcal{I}^{\{A,H\}}, \mathcal{S}, \bm{\mathcal{A}}, \mathcal{P}, \bm{\Omega}, \bm{O}, R, \gamma\rangle$, where $\mathcal{I}^{\{A,H\}}$ indicates the set of $N_A + N_H$ players, $A=\{1,\dots,N_A\}$ and $H=\{1,\dots,N_H\}$ denote the AI and human set, respectively. In our heterogeneous settings, agents and humans may have their own observations and actions, and we denote their corresponding spaces with superscripts. We consider the scenario where there is one controlled agent in each set in this paper, i.e., $N_A = N_H = 1$.  The expected discounted return can be defined as $J(\pi_A, \pi_H) = \mathbb{E}\left[\sum_{t=0}^\infty \gamma^t R(s_t, a_t^{(A)}, a_t^{(H)})\right]$, where $a_t^{(A)} \sim \pi_A  (\cdot|s_t), a_t^{(H)} \sim \pi_H  (\cdot|s_t), s_{t+1} \sim \mathcal{P}(\cdot|s_t, \{a_t^{  (A)}, a_t^{  (H)}\})$. The goal is to specify $\pi_A$ and $\pi_H$ to achieve the highest $J(\pi_A, \pi_H)$. Let $\Pi_H$ denotes the unknown distribution of human policies. The final objective of ZSC is to obtain the best AI agent policy $\pi^*_A$ with the goal of maximizing the expected return with the true but unknown partner distribution, i.e., $\pi^*_A= \arg\max_{\pi_A}\mathbb{E}_{\pi_H \sim \Pi^*_H}\left[J(\pi_A, \pi_H)\right]$.

As a common approach in competitive MARL, self-play (SP)~\cite{sp,sp2} has been used to solve the two-player ZSC task. However, the agent in self-play is only paired with a copy of itself, making it hard to generalize well to novel partners~\cite{overcooked,ZSC}. Population play (PP)~\cite{pp} maintains a population of agents that interact with each other. As there is no explicit strategy to maintain diversity, PP produces agents similar to self-play, still failing to generate robust agents for novel partners~\cite{fcp}. There are lots efforts to solve the ZSC tasks, which we will introduce in the following related work section.

\section{Related work}\label{relatedwork}

\subsection{Zero-Shot Coordination}
One of the key idea for ZSC is to expose the agents to diverse partners during the training process~\cite{TrajeDi,fcp,MEP}. Trajectory Diversity (TrajeDi)~\cite{TrajeDi} is the first to apply diversity to solve the ZSC problem, which trains a diverse population of partners with a common best-response agent. Different from this, the Fictitious Co-Play (FCP)~\cite{fcp} and Maximum Entropy Population-based training (MEP)~\cite{MEP} methods first generate a diverse population of partners, and then train a robust agent to coordinate well with these partners. 

Recently, there are also lots of methods consider improving ZSC ability from different perspectives. 
Any-play~\cite{new-zsc-anyplay} extend ZSC evaluation metrics and augment ZSC by considering the partners are trained by different algorithms. Besides, it extends diversity-based intrinsic rewards from DIAYN~\cite{DIAYN} for ZSC. 
Hidden-Utility Self-Play (HSP)~\cite{new-zsc-hsp} considers the scenario that human partners are biased according to their own preferences, modeling human biases as hidden reward functions, which can be seen as a special case of heterogeneous coordination. However, HSP does not consider the heterogeneous advantages or skills of agent and partner as what we do. 
LIPO~\cite{new-zsc-lipo} learns diverse behaviors using information about the task’s objective via the expected return (i.e., policy compatibility). 
PECAN~\cite{new-zsc-pecan} uses policy ensemble to increase the diversity of partners in the population. Besides, it develops a context-aware method enabling the agent to analyze and identify the partner’s potential policy primitives so that it can take different actions accordingly. 
Hierarchical Population Training (HiPT)~\cite{new-zsc-hierarchical} learns a hierarchical policy to further strengthen the ability of the agent. It learns multiple best-response policies as its low-level policy and learns a high-level policy that acts as a manager to control them. 
COLE~\cite{new-zsc-cole} constructs open-ended objectives from the perspective of graph theory to efficiently evaluate and identify cooperative incompatibility. 
SAMA~\cite{new-zsc-sama} uses large language models to suggest potential goals and provide suitable goal decomposition for coordination tasks, thus significantly improving the performance.

The previous methods mentioned above involve individual-based or population-based self-play, where a policy may take on different roles, which is, however, not helpful to the coordination under heterogeneity. 
Note that the heterogeneity between AI and humans is natural and fixed in many real-world scenarios. To the best of our knowledge, there is no existing work that trains ZSC agents while considering heterogeneity, i.e., different advantages or skills between AI agent and human partner. 
In this paper, our proposed MAZE method explicitly maintains two populations of agents and partners, and coevolves them by pairing agents and partners. That is, the agent and partner are fixed to be two specific players, respectively, and will not be changed throughout the training and testing process. Thus, MAZE can be more suitable for heterogeneous ZSC. 
Furthermore, as a general framework, MAZE is compatible with these methods, and their techniques can be leveraged by MAZE to further enhance the coordination capabilities in heterogeneous settings. Our work represents an initial step, introducing a relatively simple yet fundamental framework for heterogeneous ZSC. 

\subsection{Diversity in RL}
A series of works have explored the benefit of diversity in RL. For example, diversity has been found helpful in fast adaptation~\cite{animal,robot-few-shot-qd}, robust training~\cite{yuan-fcs,romance}, sparse or deceptive reward~\cite{NSR-ES,one-solution}, open-ended learning~\cite{poet,xland}, unsupervised environment designing~\cite{parkerholder2022evolving,qd4ge} and ZSC~\cite{TrajeDi,fcp,MEP}. There are various definitions of diversity~\cite{NS,DvD,TrajeDi}, and they are summarized in Liu \textit{et al.}~\cite{diversity_definition1}.

A simple way to achieve diversity is random initialization~\cite{fcp}. However, the population diversity may be gradually decreased as the algorithms continue to run. To address this issue, some algorithms add a diversity term into the objective function, such as novelty search~\cite{NSR-ES}, DIAYN~\cite{DIAYN}, Ridge Rider~\cite{ridge}, TrajeDi~\cite{TrajeDi}, etc. Quality-Diversity algorithms~\cite{QD-Framework,QD-optimization} further introduce an archive to maintain the diverse policies generated so far, and the policies in the population are iteratively selected from the archive and updated. The proposed MAZE method will also use these strategies to maintain the diversity of the two populations of agents and partners.

\subsection{Coevolution}
Evolutionary algorithms~\cite{eabook} are general-purpose heuristic optimization algorithms that maintain a population (i.e., a set of solutions) and improve it by simulating the natural evolution process, i.e., iterative reproduction and selection. 
Instead of maintaining one single population like traditional evolutionary algorithms, coevolutionary algorithms~\cite{DBLP:journals/ec/PotterJ00} maintain multiple populations and coevolve them competitively or cooperatively, which have been shown excellent performance in many scenarios~\cite{ccea-survey}. For example, Minimal Criterion Coevolution (MCC)~\cite{mcc} coevolves a population of problems (i.e., challenging environments) with a population of solutions. Similarly, Paired Open-Ended Trailblazer (POET)~\cite{poet,epoet} generates an endless progression of diverse and increasingly challenging environments while explicitly optimizing their solutions. 
Different from them, heterogeneous ZSC aims to find an agent that can coordinate well with unseen human partners, where the partners used for training can not be obtained independently by self-play. Thus, the proposed MAZE for heterogeneous ZSC obtains  coevolves the two populations of agents and partners in a cooperative way.

\section{MAZE Method}\label{method}
In this section, we introduce the proposed method, Multi-Agent Zero-shot coordination by coEvolution (MAZE). We consider the two-player heterogeneous ZSC problem, where the trained agent should coordinate with the partner unseen during training, and the agent together with the partner may have different advantages or skills. Exposing the agent to diverse partners during the training process is an effective way to improve the ability to coordinate with unseen partners. However, previous methods~\cite{TrajeDi,fcp,MEP} mainly generate the diverse partner population by self-play, which may not capture the cooperation behaviors between AI and humans well, and thus may not be suitable for heterogeneous ZSC. 

\begin{figure}[t!]
    \centering
    \includegraphics[width=0.49\textwidth]{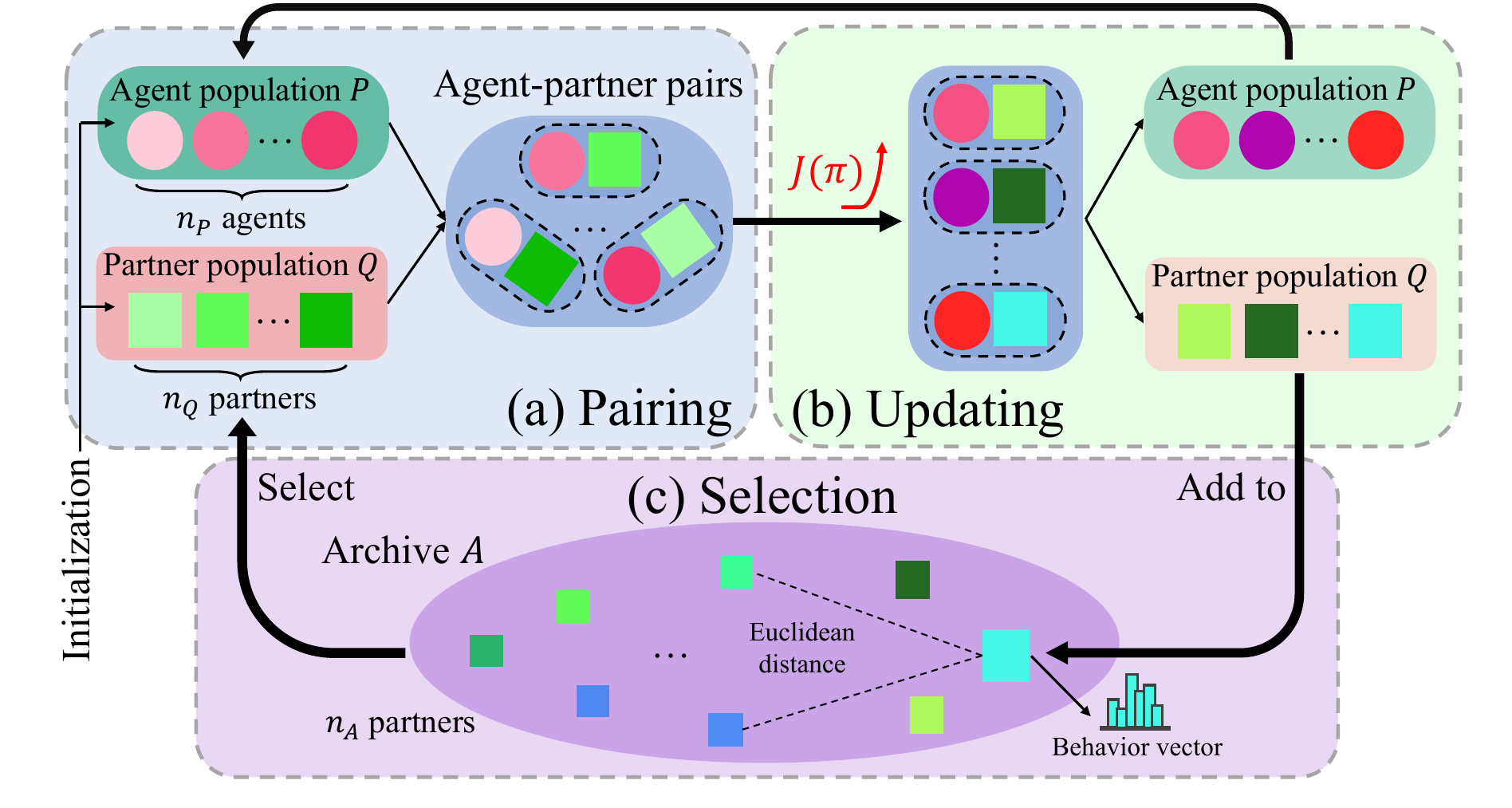}
    \caption{Illustration of the MAZE method, where MAZE coevolves two populations of agents and partners through three sub-processes, i.e., pairing, updating and selection.}
    \label{fig:maze}
\end{figure}

The detailed procedure of MAZE is presented in Algorithm~\ref{algorithm}. Instead of using population-based self-play, MAZE initializes two populations of agents and partners randomly (line~1 of Algorithm~\ref{algorithm}), and coevolves them through three sub-processes: pairing, updating and selection (lines~3--15). In each generation of MAZE, the agents and partners from the two populations are first paired to get the agent-partner pairs (lines~4--5). In the updating process (lines~6--11), each agent-partner pair interacts with the environment to collect their trajectories, which are used to update the agent and partner by optimizing an objective function with both reward and diversity. In the selection process (lines~12--13), the updated agents directly form the agent population in the next generation, while the updated partners will be added into an archive, which contains diverse partners generated so far. The next partner population is generated by selecting diverse partners from the archive instead of using the updated partners directly, with the goal of enhancing its diversity. When the training phase is finished, we deploy the final agent population by some specific strategy. The procedure of MAZE is also illustrated in Figure~\ref{fig:maze} for clarity.

\begin{algorithm}[tb]
\caption{MAZE}\label{algorithm}
\textbf{Input}: size $n_P$ of the agent population $P$, size $n_Q$ of the partner population $Q$, number $T$ of generations, number $T'$ of updating iterations in each generation, size $n_A$ of the archive $A$, and pairing strategy \emph{Pair}
\begin{algorithmic}[1] 
    \STATE Randomly initialize the agent population $P$ and the partner population $Q$;
    \STATE $t=0$ and archive $A=Q$;
    \WHILE{$t < T$}
        \STATE For each agent $\pi_{\bm\theta} \in P$, select a partner $\pi_{\bm\phi}$ from $Q$ according to the strategy \emph{Pair};
        \STATE Let $\{(\pi^{(i)}_{\bm\theta},\pi^{(i)}_{\bm\phi})\}_{i=1}^{n_P}$ denote the obtained agent-partner pairs;
        \FOR{$j=1:T'$}
            \FOR{$i=1:n_P$}
                \STATE Use $(\pi^{(i)}_{\bm\theta},\pi^{(i)}_{\bm\phi})$ to interact with the environment and collect the trajectories;
                \STATE Update the agent $\pi^{(i)}_{\bm\theta}$ and partner $\pi^{(i)}_{\bm\phi}$ by applying PPO to optimize Eq.~(\refeq{eq:loss function})
            \ENDFOR
        \ENDFOR
        \STATE Use the $n_P$ updated agents to form the next agent population $P$ directly;
        \STATE Add the updated partners into the archive $A$, and select $n_Q$ diverse partners from $A$ to form the next partner population $Q$;
        \STATE $t=t+1$
    \ENDWHILE
\end{algorithmic}
\end{algorithm}

\subsection{Pairing}\label{sec3.1 pairing}
Pairing is to select a partner from the partner population $Q$ for each agent in the agent population $P$ (line~4), forming the $n_P$ agent-partner pairs (line~5), which will be used to collect the trajectories to train the policies in the two-player environment. A straightforward pairing strategy is to pair each agent with the same partner in each generation of training. However, this fixed pairing strategy may form some conventions which may not be understood by novel partners. To break these conventions and achieve ZSC, each agent should be paired with as many different partners as possible during the training phase. Hence, it is encouraged to pair an agent with different partners of different generations. 

In this paper, the following pairing strategies are considered. 1) \emph{Fixed pairing.} An agent is paired with the same partner throughout the training process. 2) \emph{Random pairing.} An agent is paired with a partner randomly selected from $Q$ in each generation. 3) \emph{Best (Worst) pairing.} Before pairing, the agents in $P$ are evaluated with all the candidate partners in $Q$. Then, each agent is paired with the best (worst) partner it coordinates with. MAZE uses the random pairing strategy by default. But we also provide the comparison of MAZE using these different pairing strategies in the experiments. Note that there can be more complex pairing strategies, e.g., an agent can be paired with more than one partner in each iteration, which will be studied in the future. 

\subsection{Updating}\label{sec3.2 updating}
After pairing, we collect the trajectories of each agent-partner pair in line~8, and use them to update the corresponding agent and partner in line~9. When updating, we want the agent to play well with the partner; meanwhile, we should ensure the diversity of both agent and partner populations. Inspired by maximum entropy RL~\cite{sac,MEP}, we add the diversity term into the objective function and optimize it by Proximal Policy Optimization (PPO)~\cite{ppo}. Here, we use Jensen-Shannon Divergence (JSD) to measure the diversity of a policy population $\{\pi^{(i)}\}_{i=1}^{n}$, which can be calculated as
\begin{equation}\label{eq-jsd}
\begin{aligned}
    & \text{JSD}(\{\pi^{(i)}(\cdot\mid s_t)\}_{i=1}^{n}) =  \\ & 
    \frac{1}{n} \sum^n_{i=1} \sum_{a_t\in \mathcal A_{\pi}} \pi^{(i)}(a_t\mid s_t)\log\frac{\pi^{(i)}(a_t\mid s_t)}{\bar{\pi}(a_t\mid s_t)},
\end{aligned}
\end{equation}
where $\mathcal A_{\pi}$ denotes the common action space of the policies, and $\bar{\pi}(\cdot \mid s_t)=\frac{1}{n}\sum^{n}_{i=1}\pi^{(i)}(\cdot\mid s_t)$. Based on the JSD measure Eq.~(\refeq{eq-jsd}), we use policy $\pi$ to represent an agent or a partner, whose objective function can be defined as
\begin{equation}\label{eq:loss function}
    J(\pi) = \mathbb E_{\tau \sim \pi} \left[\sum_{t} R(s_t,\boldsymbol{a}_t) + \alpha \cdot  \text{JSD}(\{\pi^{(i)}(\cdot\mid s_t)\}_{i=1}^{n})\right],
\end{equation}
where $\boldsymbol{a}_t$ is the joint-action of agent and partner, $\alpha \geq 0$ is the hyper-parameter that controls the importance of the diversity term, and $\{\pi^{(i)}\}_{i=1}^{n}$ is the corresponding policy population containing $\pi$. In each generation of MAZE, each agent-partner pair will be updated for $T'$ iterations (line~6). 

\subsection{Selection}\label{sec3.3 selection}

After updating, the $n_{P}$ updated agents form the agent population $P$ in the next generation directly, as shown in line~12 of Algorithm~\ref{algorithm}. But the generation of the next partner population $Q$ in line~13 is more complicated. In the ZSC setting, the trained agent may coordinate with any unseen partner, so we want to expose the agent to the partners with different skill levels during the training phase. Note that the partners with different skill levels can be simulated by the partners in different training stages~\cite{fcp}. To achieve this goal, we keep an archive $A$, which stores the diverse partners encountered during the training process. In each generation, we add the updated partners into the archive $A$ and select $n_Q$ diverse partners from $A$, forming the partner population $Q$ in the next generation.

To be specific, MAZE maintains an archive $A$ with capacity $n_A$. After updating, the updated partners and those partners already in $A$ will be evaluated together with the $n_{P}$ updated agents to get the pair-wise performance. That is, each partner will be associated with an $n_P$-dim vector, each dimension of which is the reward achieved by the partner when cooperating with a corresponding agent. This $n_P$-dim vector will be used to represent the behavior of the partner. Then, the distance between two partners is measured by the Euclidean distance between their corresponding behavior vectors. We add the updated partners into the archive $A$ one by one. Whenever we want to add a new policy (i.e., partner) into $A$, the policy with the most similar behavior is first selected from $A$. If their behavior distance is greater than a threshold, we add the new policy into $A$. Otherwise, we randomly keep one. Note that when the archive size $|A|$ exceeds the capacity $n_A$ after adding a new policy, the oldest policy in the archive will be deleted. After that, we select $n_Q$ diverse partners from the updated $A$ to form the next partner population $Q$. Specifically, inspired by EDO-CS~\cite{EDO-CS}, we divide $A$ into $n_Q$ clusters by $K$-means based on the behavior vector and randomly select one good partner from each cluster.

\subsection{Deployment}\label{sec3.4 deployment}
After running $T$ generations, MAZE obtains a population of diverse agents, instead of only a single best-response agent. There are two natural ways to deploy the obtained agent population. One is the ensemble of the agents by majority voting, which selects the action voted by the most agents given a state. The other is choosing the best response agent by offline evaluation (BRO) with the partners in the archive $A$. That is, each agent is evaluated with all the partners in the archive $A$, and the agent achieving the highest average reward will be selected. These two deployment strategies are briefly called Ensemble and BRO, respectively. MAZE uses the Ensemble strategy by default. Its performance of using these two deployment strategies will be compared in the experiments.
 
Finally, we want to summarize the main idea of MAZE. As the previous methods using self-play may not capture the cooperation behaviors between AI and humans well in heterogeneous settings, MAZE uses two different policies to represent the agent and partner, respectively. The simplest implementation is to train them directly without changing training partners, called Vanilla-MAZE (V-MAZE). That is, V-MAZE simply takes heterogeneity into account. In fact, V-MAZE has already performed well on heterogeneous tasks, which will be shown in RQ1 of experiments. Inspired by previous works~\cite{pp,TrajeDi,MEP,QD-Framework}, MAZE further tries to improve the diversity of partners to obtain better ZSC agents through three different ways: 1) maintaining a population and changing the paired policies in the sub-process of pairing; 2) adding diversity terms into the objective function rather than just maximizing rewards in the sub-process of updating; 3) actively selecting diverse partners from an archive for the next generation in the sub-process of selection. To verify the necessity and effectiveness of the above-proposed components, we will conduct ablation studies, starting from the \emph{simplest} V-MAZE and adding these components gradually until the \emph{complete} MAZE, which will be shown in RQ3 of experiments.

\section{Experiments}\label{sec4 experiemnt}

In this section, we first introduce the environments used for our experiments and the experimental settings. Then, we answer the following four research questions (RQs): 1) Is the heterogeneity worth considering? 2) How does MAZE perform compared with recently proposed algorithms? 3) How do different components of MAZE influence its performance? 4) Can MAZE coordinate well with real human partners? Finally, we provide some additional results, including the influence of larger network, visualization of partner diversity, running time analysis of MAZE, and comparison between homogeneous and heterogeneous environments. 
Our code is available at \url{https://github.com/lamda-bbo/maze}.

\subsection{Environments}

\begin{figure}[htbp]
    \centering
    \includegraphics[width=0.98\linewidth]{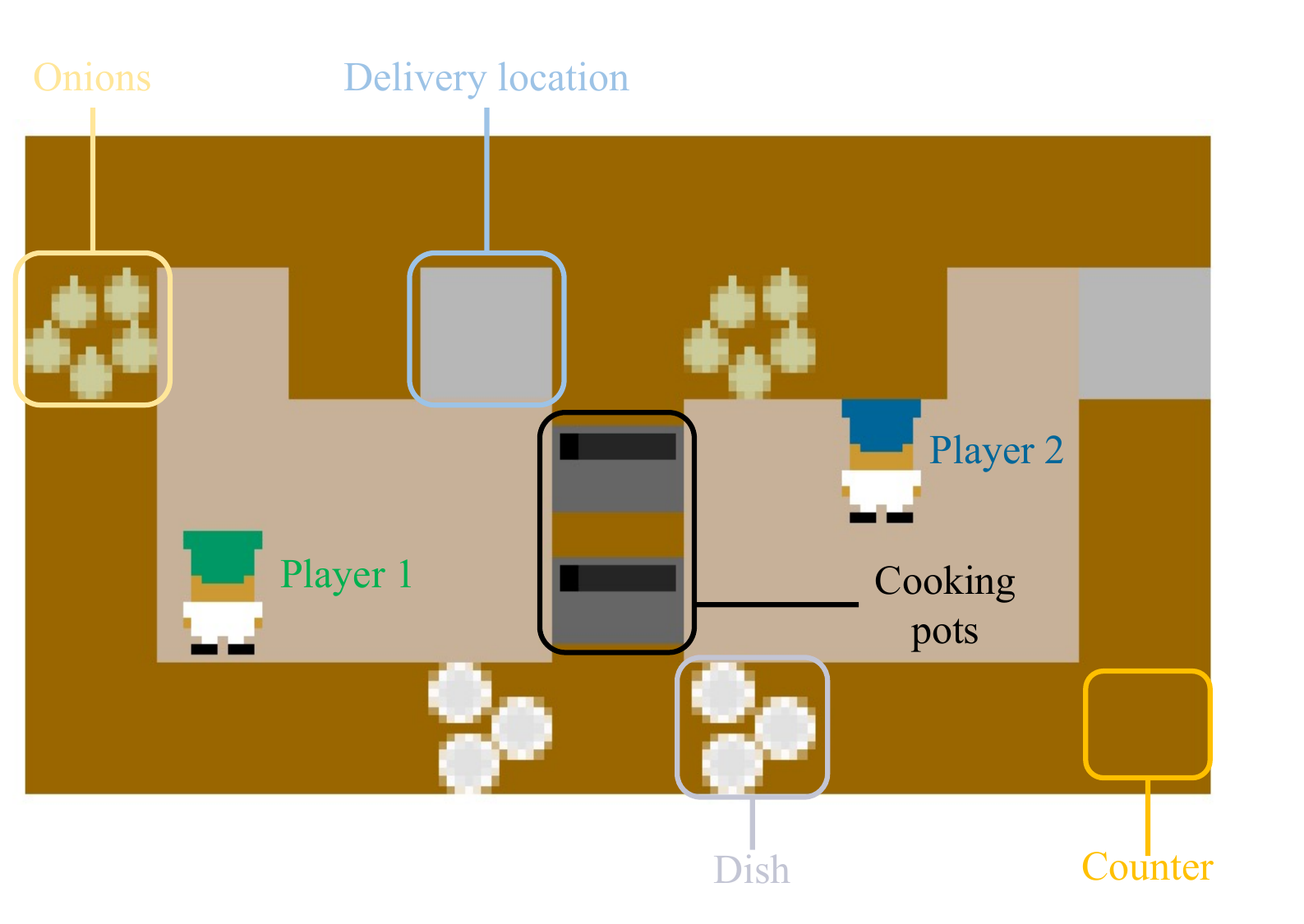}
    \caption{The \emph{AA} layout of \emph{Overcooked} environment . The \emph{Overcooked} environment is a two-player common-payoff game, where players need to coordinate to cook and deliver soup.}
    \label{fig:overcook env}
\end{figure}

To examine the performance of MAZE, we mainly conduct experiments on the \emph{Overcooked} environment~\cite{overcooked}, which is one of the most popular benchmark in ZSC~\cite{MEP,new-zsc-lipo,new-zsc-hsp,new-zsc-cole,new-zsc-pecan}.  As shown in Figure~\ref{fig:overcook env}, two players are placed into a grid-world kitchen as chefs and tasked with delivering as many cooked onions soup as possible within a limited time budget. This task involves a series of sequential high-level actions to which both players can contribute: collecting \emph{onions}, depositing them into \emph{cooking pots} and letting the onions cook into soup, collecting a \emph{dish} and getting the soup, and delivering the soup at the \emph{delivery location}. Both players are rewarded equally when delivering a soup successfully. 

\begin{figure*}[t!]
    \centering
    \includegraphics[height=2.6cm]{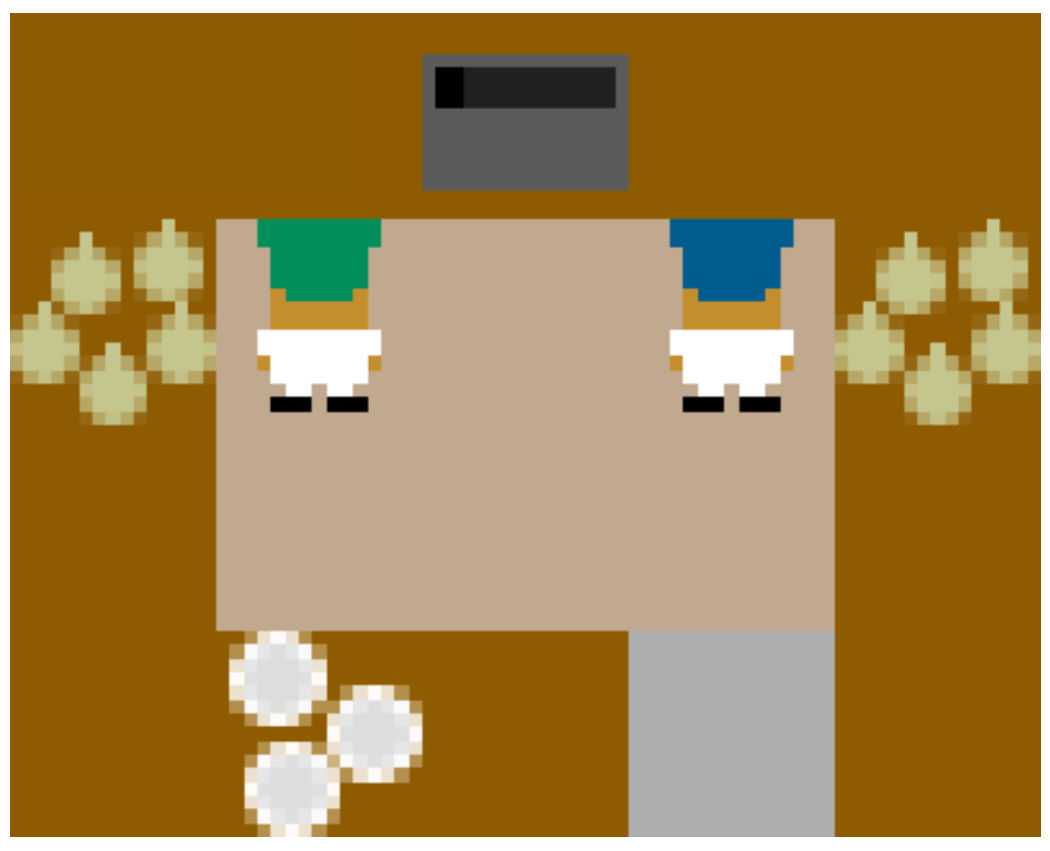} 
    \includegraphics[height=2.6cm]{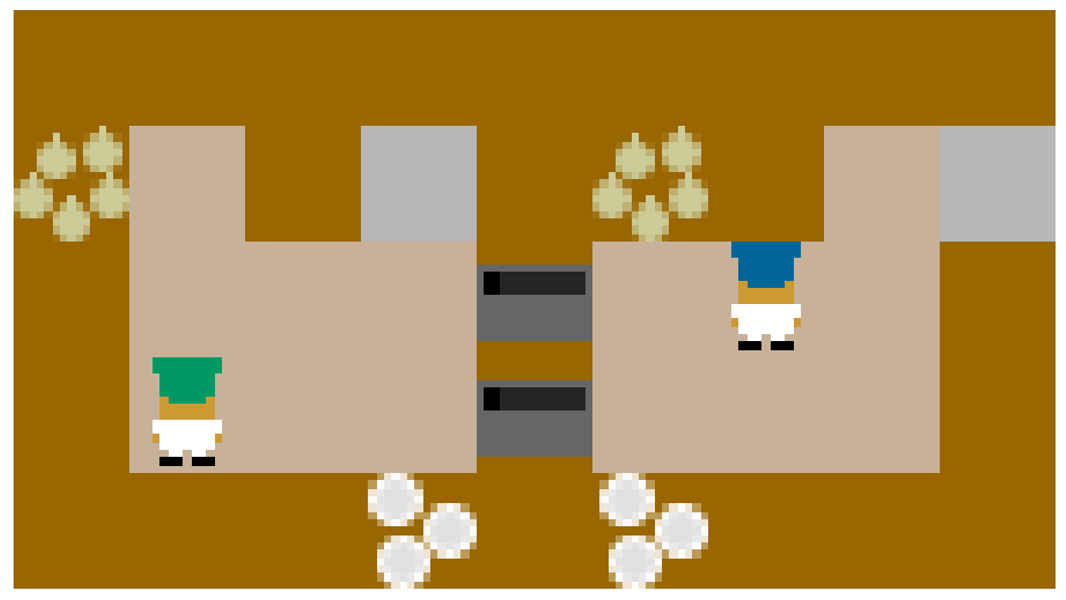} 
    \includegraphics[height=2.6cm]{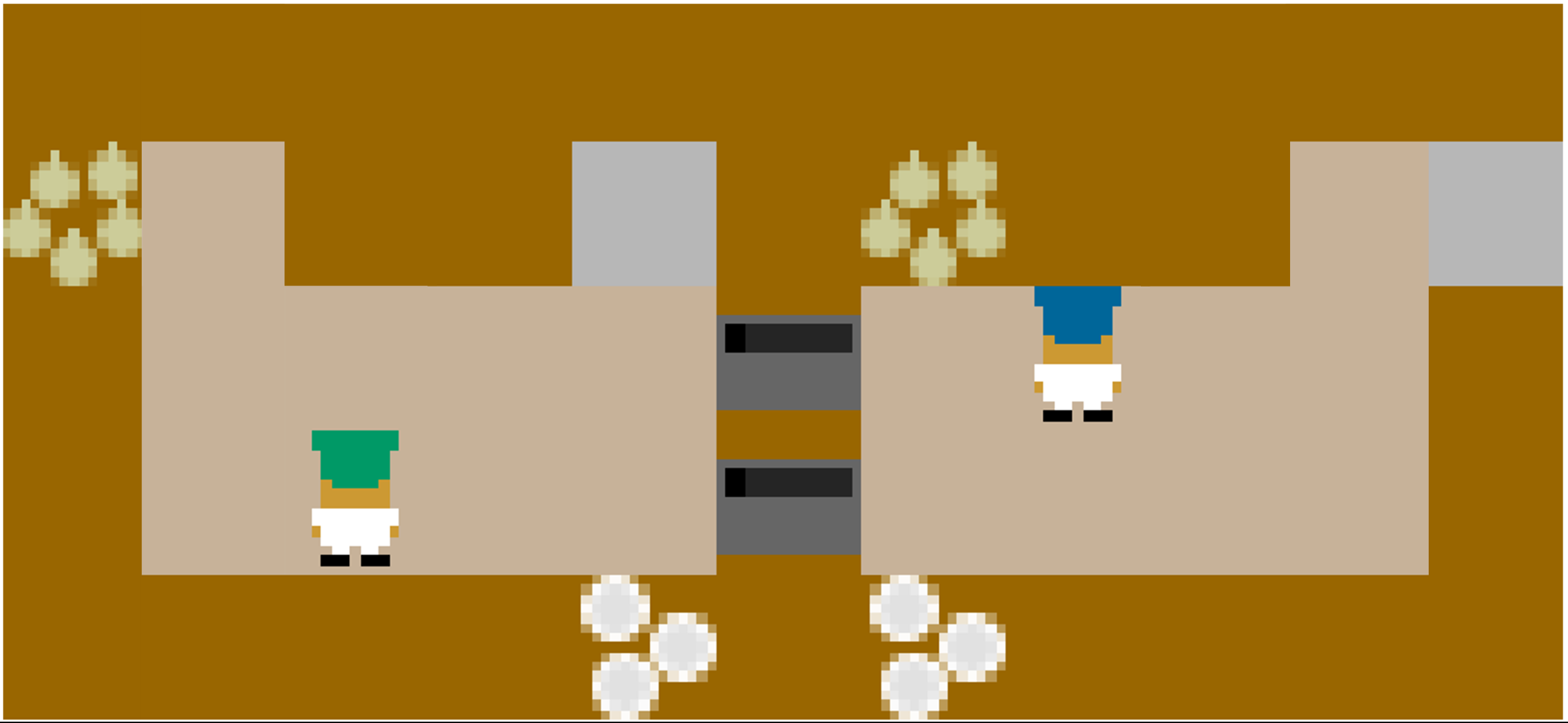} 
    \includegraphics[height=2.6cm]{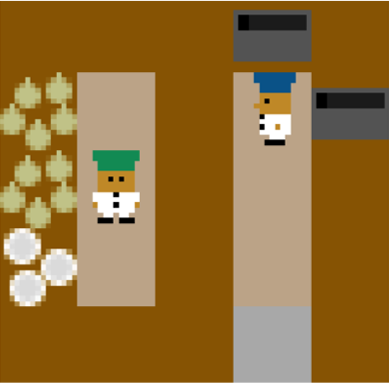}  \\
    \begin{minipage}[c]{0.23\linewidth}\centering
    \small (a) \emph{H-CR}
    \end{minipage} 
    \begin{minipage}[c]{0.24\linewidth}\centering
    \small (b) \emph{AA}
    \end{minipage}
    \begin{minipage}[c]{0.30\linewidth}\centering
    \small (c) \emph{AA-2}
    \end{minipage}
    \begin{minipage}[c]{0.19\linewidth}\centering
    \small (d) \emph{FC} \\
    \end{minipage}
    \caption{Illustration of different layouts on Overcooked.}
    \label{fig:overcooked_layouts}
\end{figure*}

Concretely, four layouts with different levels of heterogeneity, i.e., \emph{H-CR}, \emph{AA}, \emph{AA-2}, and \emph{FC}, are considered, as shown in Figure~\ref{fig:overcooked_layouts}. 
\begin{itemize}
    \item The two players in Heterogeneous-Cramped Room (H-CR) are in the same room. However, their skills are different. Specifically, player 1 can only collect \emph{onions}, deposit them into \emph{cooking pots}, but can not collect a \emph{dish}, get the soup, or deliver the soup at the \emph{delivery location}. Thus, HCR forcing them to complete the task cooperatively. 
    \item \emph{Asymmetric Advantages (AA)} is a typical heterogeneous layout , where the players in different rooms (i.e., left and right) have different advantages. Player 1 on the left is good at collecting dishes and delivering the soup, while player 2 on the right is good at collecting onions and deposing them into cooking pots. If the two players can leverage their respective advantages and coordinate well, higher scores can be achieved.
    \item Different from \emph{AA}, the players in \emph{AA-2} have their own advantages more obviously. Take player 1 as an example. The distance between the onions and the cooking pots in \emph{AA-2} is longer compared to \emph{AA}, implying that player 1 will take more time to get the onions and depose them into cooking pots. Thus, player 2's skill advantage in collecting onions and deposing them into cooking pots will be more obvious than player 1.
    \item \emph{Forced Coordination (FC)} is a more heterogeneous layout in the \emph{Overcooked} environment. Player 1 on the left can only collect and deliver the onions and dishes, while player 2 on the right is only to receive the onions, put them into the cooking pots, and then use the dish to deliver the soup. Similar to \emph{HCR}, the players in \emph{FC} \emph{must} coordinate with each other to successfully cook and deliver the soup. 
\end{itemize}

While \emph{Overcooked} is one of the most popular environments for ZSC, additional experiments on different environments could further strengthen our findings. To that end, we design a new environment called \emph{FillInTheGrid}, a simple grid-based game to verify the versatility of our framework.

\begin{figure}[htbp]
    \centering
    \includegraphics[width=0.47\linewidth]{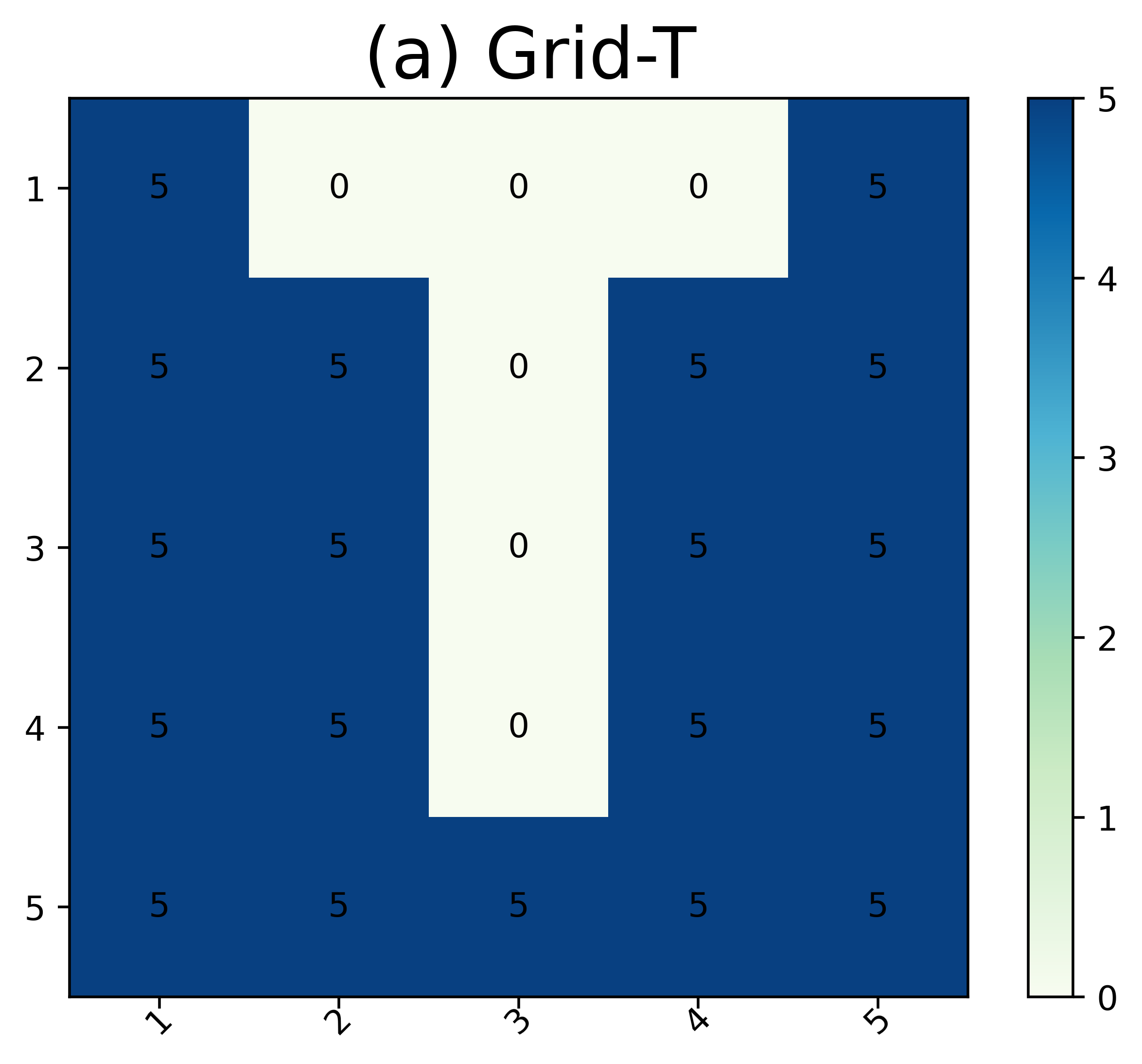}
    \includegraphics[width=0.47\linewidth]{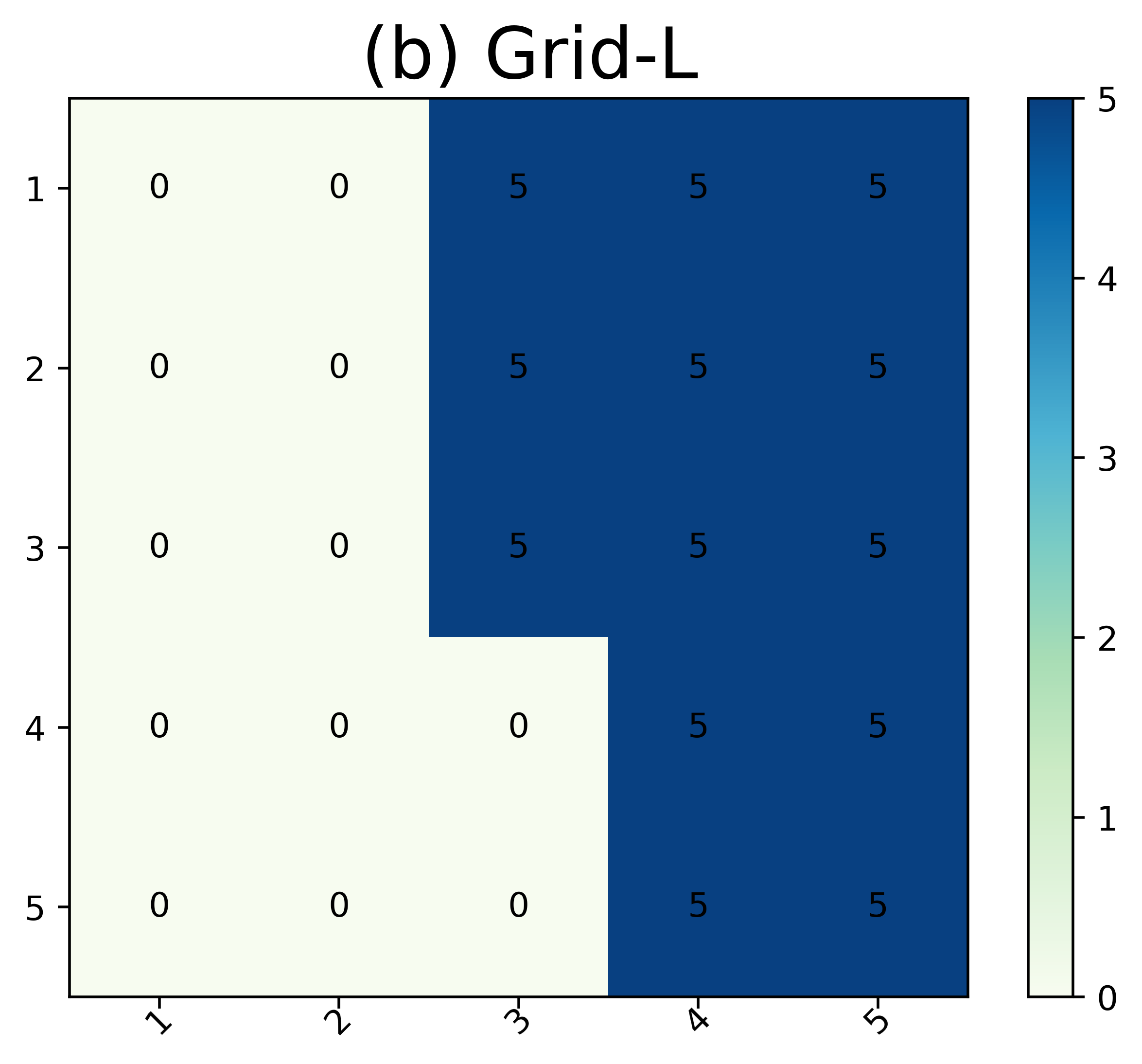} 
    \caption{\emph{FillInTheGrid} environments. (a) \emph{Grid-T} layout. (b) \emph{Grid-L} layout.} 
    \label{fig}
\end{figure} 

As shown in Figure~\ref{fig}, the game consists of a grid of cells with different values. The goal of \emph{FillInTheGrid} is to fill in the entire grid, to make all cells reach the maximum value (which is 5 in our experiments). Player 1 and player 2 share the same state space (i.e., the current grid), and each player's action is to select a cell. They have heterogeneous abilities. When player 1 selects a cell, the value of the cell as well as its left and right neighbor cells will increase by 1. When player 2 selects a cell, the value of the cell as well as its upper and lower neighbor cells will increase by 1. For example, if player 1 selects cell (2,1), the values in cells (1,1), (2,1), and (3,1) will increase by 1. Note that once a cell has reached the maximum value 5, it will be kept. The reward is defined as the value added to the grid. The episode ends when the grid is filled full of 5. To encourage the algorithm to fill cells as quickly as possible, we provide a bonus based on the number of steps that have been used. 
Compared with \emph{Overcooked}, \emph{FillInTheGrid} is a more simpler environment for investigating the heterogeneous coordination ability. Thus, we compare the training process of V-MAZE, SP, and PP to clearly show the advantages of considering heterogeneity.

\subsection{Experimental settings}
The hyper-parameters used in the experiments are summarized as follows. We summarize the detailed settings of environments, policy, and training process in Table~\ref{tab:exp-settigns}.

\begin{table}[h]
\centering
\caption{Detailed experimental settings.}
\label{tab:exp-settigns}
\begin{tabular}{ccc}
\toprule
Types        & Setting                                     & Value \\
\midrule
\multirow{3}{*}{Environments} & Steps per agent                             & 7e6   \\
             & Horizon                                     & 400   \\
             & Number of parallel rollouts                 & 50    \\
             \midrule
\multirow{4}{*}{Policy}      & Number of convolution layers                & 3     \\
             & Number of filters  & 25    \\
             & Number of hidden layers                     & 3     \\
             & Size of hidden layers                       & 64    \\
             \midrule
\multirow{6}{*}{Training}     & PPO learning rate                           & 8e-4  \\
             & PPO clipping factor                         & 0.05  \\
             & Reward shaping horizon                      & 5e6   \\
             & Discounting factor                          & 0.99  \\
             & Value function coefficient                  & 0.5   \\
             & Advantage discounting factor                & 0.98 \\ \bottomrule
\end{tabular}
\end{table}

For each method, we introduce its own implementation and hyperparameters as follows. 
For a fair comparison, the hyper-parameters of the compared algorithms in the following experiments are set as same as possible, where the population size is always set to $5$.  

\fix{\paragraph{SP} is a baseline in ZSC, which makes the agent pair with a copy of itself to interact with the environment and update. We employ the commonly used implementation of SP\footnote{\url{https://github.com/HumanCompatibleAI/human_aware_rl/tree/neurips2019}}.}

\fix{\paragraph{PP} is another baseline in ZSC, which maintains a population of agents that pair with each other.  We employ the commonly used implementation of PP\footnote{\url{https://github.com/HumanCompatibleAI/human_aware_rl/tree/neurips2019}}.}

\fix{\paragraph{TrajeDi} trains a diverse population of partners by a trajectory-based diversity measurement and a common best-response agent to these partners. 
Our implementation refers to the official demo\footnote{\url{https://colab.research.google.com/drive/1yIiw0Qs_d3O_JzGxtyvxB--jJ2M0h3Ki}}, where the diversity coefficient $\alpha$ is set to $4$.}

\fix{\paragraph{FCP} first trains a set of diverse partners and then trains a robust agent to coordinate well with these partners. The diversity is achieved by different random seeds and different checkpoints (i.e., different training stages). We maintain $5$ self-play partners initialized with $5$ different seeds (i.e., 1000--5000), respectively. Besides, we save the first checkpoint (i.e., a randomly initialized partner), the middle checkpoint (i.e., a half-trained partner), and the last checkpoint (i.e., a fully trained partner) to obtain the diverse population of partners.}

\fix{\paragraph{MEP} iis similar to FCP, which also first trains a set of diverse partners and then trains a robust agent to coordinate well with these partners. The diversity is achieved by an effective entropy-based loss term. We use the reference implementation of the authors for MEP\footnote{\url{https://github.com/ruizhaogit/maximum_entropy_population_based_training}} and employ the default parameter settings (i.e., the weight of the population entropy reward is $0.04$ for the \emph{FC} layout and $0.01$ for other layouts).}

\fix{\paragraph{COLE} constructs open-ended objectives from the perspective of graph theory to efficiently evaluate and identify cooperative incompatibility. We use the reference implementation of the authors for COLE\footnote{\url{https://github.com/liyang619/COLE-Platform/tree/COLE_training}} and employ the default parameter settings and population size $5$.}

\paragraph{MAZE} Our implementation is based on the code.\footnote{\url{https://github.com/HumanCompatibleAI/human_aware_rl/tree/neurips2019}} We use the same weight of the diversity term as MEP. The $n_P$ of agent population and $n_Q$ of partner population is set to $5$, the size $n_A$ of archive is $20$, and the number $T'$ of updating iterations in each generation is $5$.

We will report the average results across five identical seeds (1000, 2000, 3000, 4000, and 5000) for all algorithms and all tasks. The experiments are conducted on a PC with AMD Ryzen 9 3900X 12-Core Processor, NVIDIA 2080Ti, and 128GB RAM. Our code is provided in the supplementary material.

\subsection{RQ1: Is the heterogeneity worth considering?}\label{rq1}

\begin{figure*}[t!]
    \centering
    \includegraphics[width=0.30\textwidth]{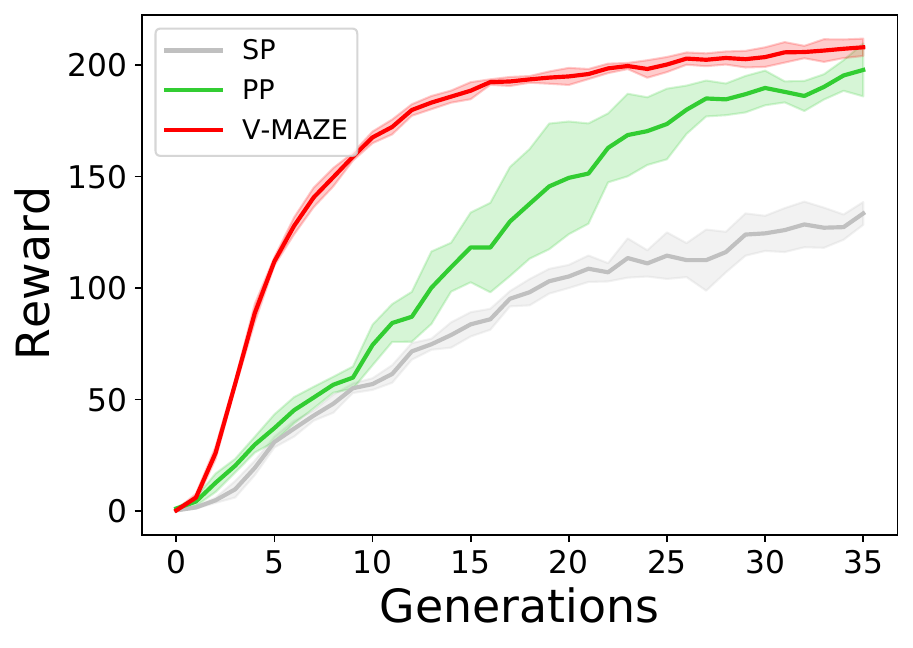} 
    \includegraphics[width=0.30\textwidth]{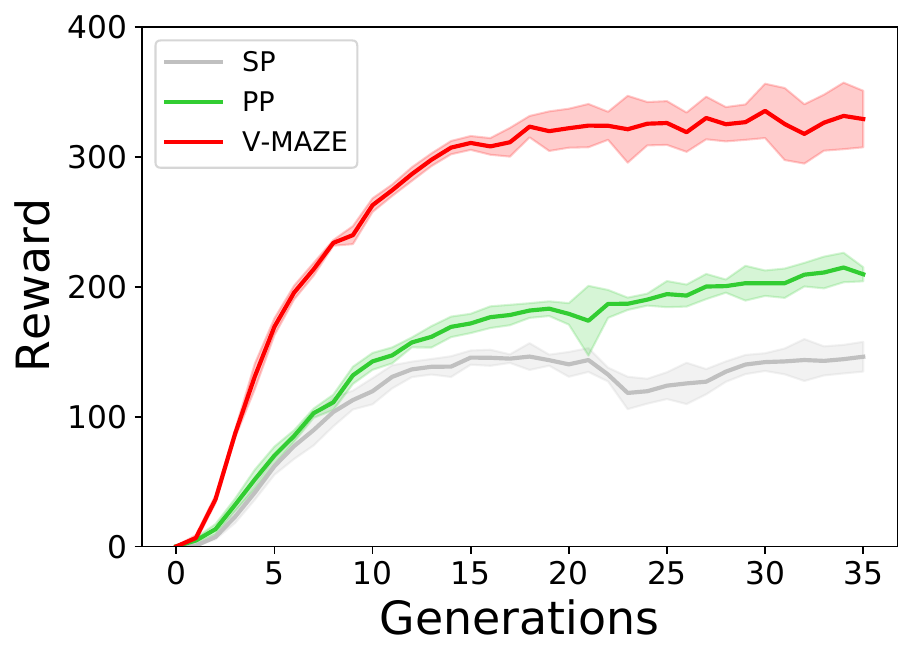}
    \includegraphics[width=0.30\textwidth]{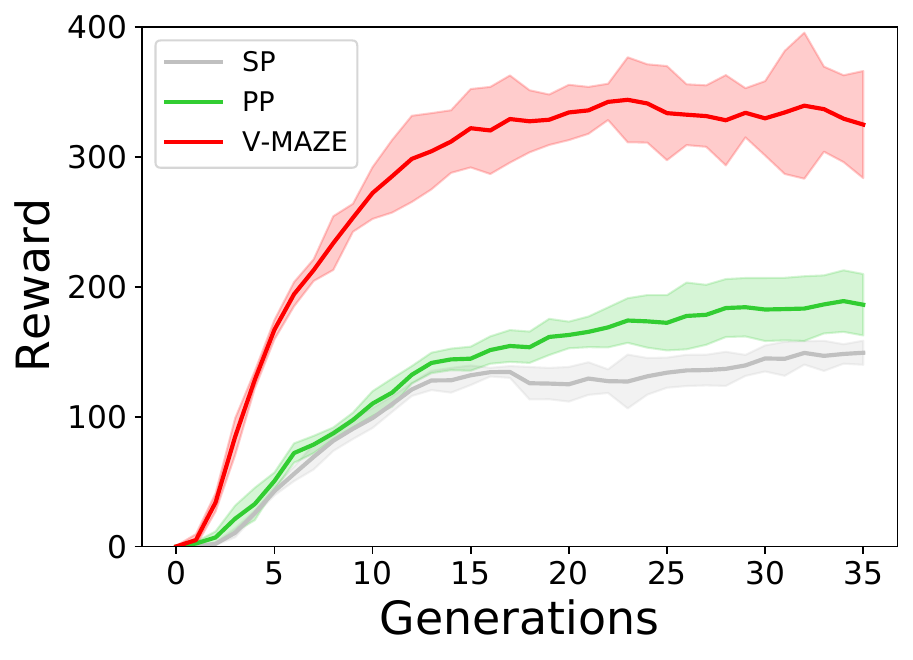}\\
    \begin{minipage}[c]{0.30\textwidth}\centering
    \small (a) \emph{H-CR}
    \end{minipage}
    \begin{minipage}[c]{0.30\textwidth}\centering
    \small (b) \emph{AA}
    \end{minipage}
    \begin{minipage}[c]{0.30\textwidth}\centering
    \small (c) \emph{AA-2}
    \end{minipage} \\
    \includegraphics[width=0.30\textwidth]{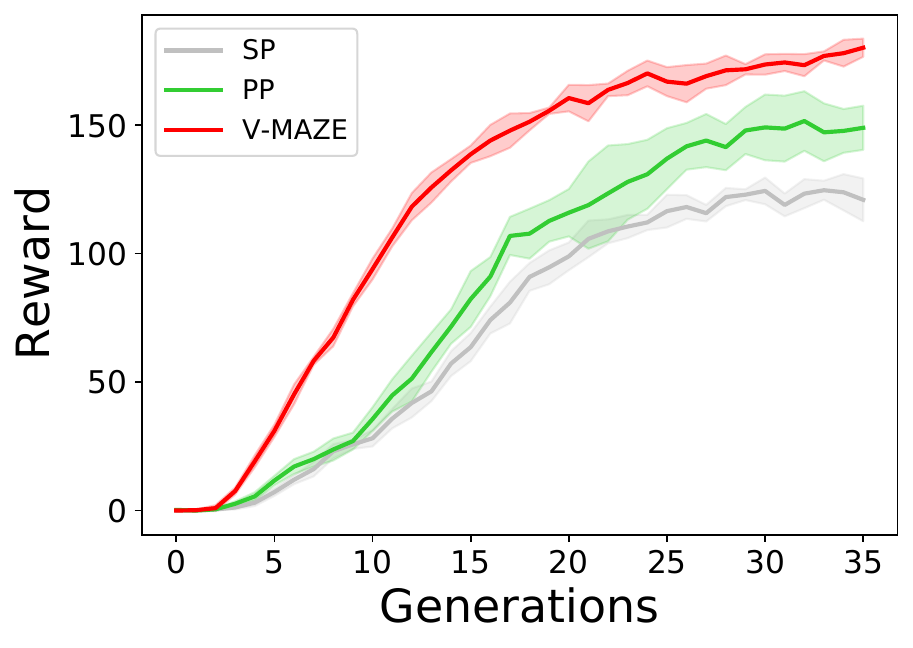}
    \includegraphics[width=0.30\textwidth]{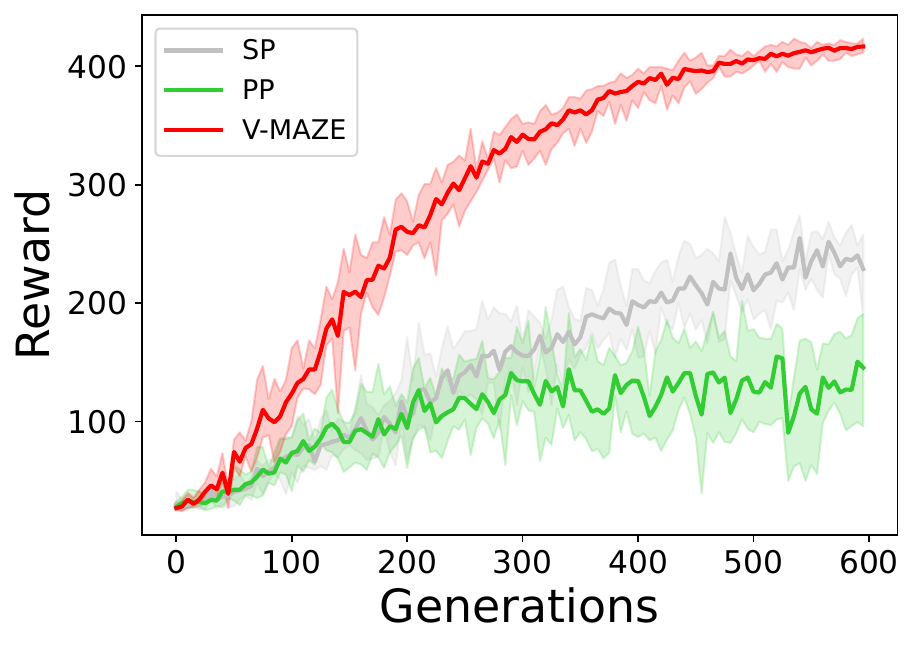}
    \includegraphics[width=0.30\textwidth]{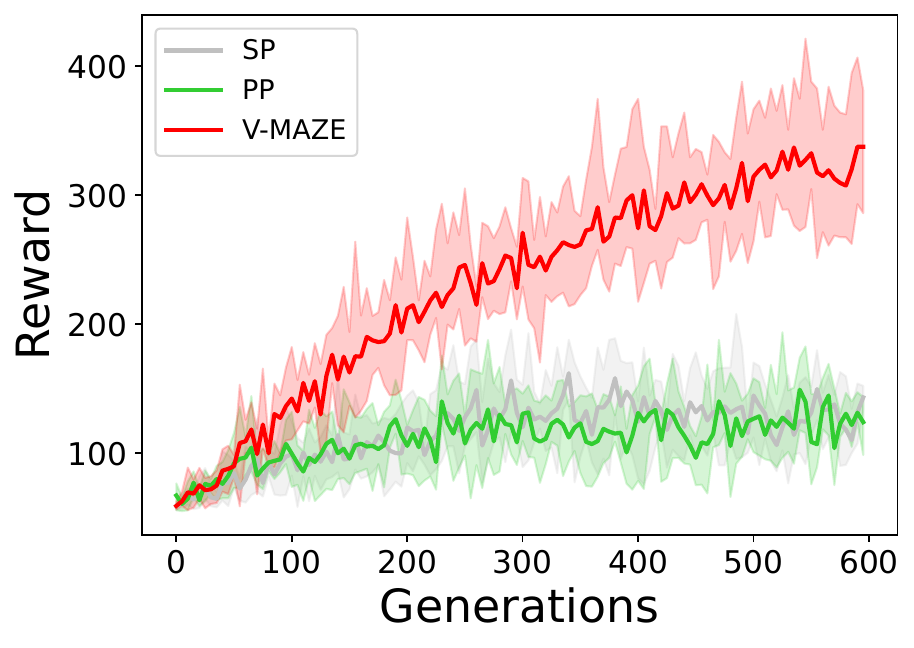}\\
    \begin{minipage}[c]{0.30\textwidth}\centering
    \small (c) \emph{FC}
    \end{minipage}
    \begin{minipage}[c]{0.30\textwidth}\centering
    \small (d) \emph{Grid-T}
    \end{minipage}
    \begin{minipage}[c]{0.30\textwidth}\centering
    \small (e) \emph{Grid-L}
    \end{minipage}
    \caption{Training curves of SP, PP and V-MAZE on four layouts from the \emph{Overcooked} environment and two layouts from the \emph{FillInTheGrid} environment.}
    \label{fig:main fig}
\end{figure*}

We first investigate if considering heterogeneity is valuable during the training process. \fix{To purely and clearly examine the importance of considering heterogeneity, we implement the \textit{simplest variant} of MAZE,} i.e., V-MAZE, which uses the fixed pairing strategy, updates the policies without considering the diversity term in the objective function, and does not use the archive for partner selection. That is, the only difference between V-MAZE and SP is that V-MAZE maintains two separate populations of agents and partners. We compare V-MAZE with SP and PP, which use self-play and single-population play, respectively. \fix{Note that we did not include the results of the methods (TrajeDi, FCP, and MEP), because they have other complex mechanisms, such as diversity-based updates to improve ZSC ability, and thus the performance difference cannot be purely attributed to the mechanism of considering heterogeneity.}

\begin{table}[t!]
\centering
\caption{\fix{The reward (mean$\pm$std.) achieved by the compared algorithms when testing with different partners on different layouts. For each combination of layout and partner, the largest reward is \textbf{bolded}.}}
\resizebox{\linewidth}{!}{
\begin{tabular}{l|c|ccc}
\toprule
Layout                     & Partner     & SP                       & PP        & V-MAZE                    \\ \midrule
\multirow{4}{*}{\emph{H-CR}} & Random      & 3.8 $\pm$ 2.1      & \textbf{5.0 $\pm$ 1.5}   & 4.0 $\pm$ 0.7         \\
                        & Self-Play   & 40.4 $\pm$ 9.5        & 36.8 $\pm$ 6.4  & \textbf{50.3 $\pm$ 7.1}            \\
                          & MAZE        & 100.0 $\pm$ 14.4    & \textbf{143.7 $\pm$ 17.1}   & 124.5 $\pm$ 14.2        \\
                         & Human Proxy & 61.4 $\pm$ 10.3      & \textbf{73.6 $\pm$ 10.3}     & 69.5 $\pm$ 10.1         \\ \midrule
\multirow{4}{*}{\emph{AA}}   & Random      & 30.6 $\pm$ 7.6        & 36.0 $\pm$ 9.1    & \textbf{48.0 $\pm$ 8.7}          \\
                        & Self-Play   & 172.6 $\pm$ 10.4   & 204.8 $\pm$ 10.8 & \textbf{207.0 $\pm$ 20.9} \\
                        & MAZE        & 213.8 $\pm$ 17.0      & 223.0 $\pm$ 10.1     & \textbf{310.5 $\pm$ 52.6}      \\
                         & Human Proxy & 38.5 $\pm$ 10.0       & \textbf{43.9 $\pm$ 2.9}   & 43.5 $\pm$ 6.0          \\ \midrule
\multirow{4}{*}{\emph{AA}-2} & Random      & 28.0 $\pm$ 6.7        & 29.8 $\pm$ 7.0  & \textbf{52.0 $\pm$ 10.6}           \\
                        & Self-Play   & 52.4 $\pm$ 10.4       & 79.6 $\pm$ 19.4  & \textbf{123.4 $\pm$ 29.9}          \\
                          & MAZE        & 132.6 $\pm$ 29.7      & 152.8 $\pm$ 24.6 & \textbf{308.2 $\pm$ 34.3}         \\
                        & Human Proxy & 39.4 $\pm$ 6.2        & 33.8 $\pm$ 9.3  & \textbf{60.4 $\pm$ 14.3}           \\ \midrule
\multirow{4}{*}{\emph{FC}}  & Random      & 3.3 $\pm$ 2.2         & 3.0 $\pm$ 0.6 & 4.0 $\pm$ 1.9              \\
                          & Self-Play   & 74.6 $\pm$ 17.6       & 76.8 $\pm$ 25.2  & \textbf{99.0 $\pm$ 37.9}         \\
                        & MAZE        & 96.5 $\pm$ 16.0       & 92.6 $\pm$ 11.6   & \textbf{108.5 $\pm$ 26.0}         \\
                           & Human Proxy & 21.3 $\pm$ 2.4  & \textbf{28.1 $\pm$ 13.3}  & 18.3 $\pm$ 5.3  \\ \midrule
\multirow{2}{*}{\emph{Grid-T}}  & Self-Play      & 79.3 $\pm$ 12.4         & 168.9 $\pm$ 17.2 & \textbf{242.4 $\pm$ 25.1}             \\
                          & MAZE   & 197.3 $\pm$ 10.9         & 215.2 $\pm$ 20.8 & \textbf{358.6} $\pm$ 19.2          \\ \midrule
\multirow{2}{*}{\emph{Grid-L}}  & Self-Play      & 74.2 $\pm$ 11.6         & 79.5 $\pm$ 14.3 & \textbf{201.0} $\pm$ 19.7               \\
                          & MAZE   & 184.8 $\pm$ 7.4         & 193.2 $\pm$ 8.9 & \textbf{304.5} $\pm$ 17.4           \\ \midrule
\multicolumn{2}{c|}{Average Rank}                             & 2.75                  & 1.95   & \textbf{1.3}
\\ \bottomrule
\end{tabular}
}\label{tab:baseline-test}
\end{table}

\begin{table*}[t!]
\centering
\caption{\fix{The reward (mean$\pm$std.) achieved by the compared algorithms when testing with different partners on different layouts. For each combination of layout and partner, the largest reward is \textbf{bolded}. The symbols `$+$', `$-$' and `$\approx$' indicate that the result
is significantly superior to, inferior to, and almost equivalent to MAZE, respectively, according to the Wilcoxon rank-sum test with significance level 0.05.}}
\resizebox{\linewidth}{!}{
\begin{tabular}{l|c|ccccc|c|c}
\toprule
     Layout          & Partner    & PP    & TrajeDi & FCP   & MEP    & COLE & MAZE \\ \midrule
\multirow{4}{*}{\emph{H-CR}} & Random   & 5.0 $\pm$ 1.5 $-$   & 5.5 $\pm$ 0.7  $-$  & 6.4 $\pm$ 1.1 $\approx$ & 6.5 $\pm$ 0.7 $\approx$  & 6.7 $\pm$ 0.6 $\approx$ & \textbf{7.0 $\pm$ 0.7}    \\
& Self-Play   & 36.8 $\pm$ 6.4 $-$ & 87.6 $\pm$ 9.0 $\approx$ & \textbf{121.6 $\pm$ 12.9} $+$   & 85.8 $\pm$ 7.5 $\approx$  & 88.9 $\pm$ 6.8 $\approx$ & 97.5 $\pm$ 14.6 \\
& MAZE     & 143.7 $\pm$ 17.1 $-$ & 153.8 $\pm$ 16.5 $-$ & 140.4 $\pm$ 15.5 $-$ & 129.8 $\pm$ 14.9 $-$ & 138.2 $\pm$ 13.6 $-$ & \textbf{183.7 $\pm$ 11.9} \\
& Human Proxy  & 73.6 $\pm$ 10.3 $-$ & 81.0 $\pm$ 13.3 $-$  & 101.2 $\pm$ 15.5 $\approx$ & 90.2 $\pm$ 10.0 $-$ & 102.4 $\pm$ 9.3 $\approx$ & \textbf{118.6 $\pm$ 13.2} \\\midrule                    
\multirow{4}{*}{\emph{AA}} & Random & 36.0 $\pm$ 9.1 $-$ & 49.3 $\pm$ 11.9 $-$  & 50.4 $\pm$ 7.4 $-$  & 63.5 $\pm$ 12.5 $\approx$  & 59.7 $\pm$ 9.3 $-$ & \textbf{74.4 $\pm$ 15.5}  \\           
                    & Self-Play   & \textbf{204.8 $\pm$ 10.8} $+$ & 180.5 $\pm$ 15.5 $+$  & 171.0 $\pm$ 11.5 $+$  & 190.3 $\pm$ 10.1 $+$ & 197.4 $\pm$ 12.6 $+$ & 142.6 $\pm$ 20.3 \\
                    & MAZE     & 223.0 $\pm$ 10.1 $-$ & 247.0 $\pm$ 27.1 $-$ & 236.0 $\pm$ 12.9 $-$ & 120.0 $\pm$ 12.4 $-$ & 278.2 $\pm$ 11.4 $-$  & \textbf{315.5 $\pm$ 10.1} \\
                    &Human Proxy   & 43.9 $\pm$ 2.9 $-$  & 50.2 $\pm$ 6.0 $-$   & 38.4 $\pm$ 4.2 $-$  & 92.0 $\pm$ 29.5 $\approx$  & 103.5 $\pm$ 14.1 $\approx$ & \textbf{112.0 $\pm$ 19.1} \\\midrule
\multirow{4}{*}{\emph{AA-2}} & Random   & 29.8 $\pm$ 7.0 $-$ & 45.5 $\pm$ 10.1 $\approx$  & 43.5 $\pm$ 10.5 $\approx$  & \textbf{58.4 $\pm$ 14.8} $\approx$  & 46.3 $\pm$ 12.1 $\approx$  & 56.5 $\pm$ 9.3   \\
  
                    & Self-Play & 79.6 $\pm$ 19.4 $-$ & 100.8 $\pm$ 12.5 $-$ & 121.2 $\pm$ 15.5 $\approx$ & 108.0 $\pm$ 29.9 $\approx$ & 125.7 $\pm$ 14.7 $\approx$ & \textbf{130.6 $\pm$ 18.5} \\
                    & MAZE    & 152.8 $\pm$ 24.6 $-$ & 170.5 $\pm$ 11.1 $-$  & 156.0 $\pm$ 15.3 $-$ & 124.6 $\pm$ 29.3 $-$ & 298.5 $\pm$ 14.8 $-$ & \textbf{381.4 $\pm$ 13.7} \\
                    & Human Proxy  & 33.8 $\pm$ 9.3 $-$ & 48.0 $\pm$ 6.9 $-$ & 55.8 $\pm$ 15.2 $-$  & 79.0 $\pm$ 11.2 $-$  & 92.3 $\pm$ 14.1 $-$  & \textbf{111.5 $\pm$ 13.7} \\\midrule
\multirow{4}{*}{\emph{FC}} & Random   & 3.0 $\pm$ 0.6 $-$  & 6.0 $\pm$ 0.7 $-$   & 5.1 $\pm$ 0.7 $-$  & 6.0 $\pm$ 1.0 $-$   & \textbf{7.6 $\pm$ 2.0} $\approx$  & 7.3 $\pm$ 0.4    \\
                    & Self-Play   & 76.8 $\pm$ 25.2 $-$  & 80.8 $\pm$ 22.4 $-$   & 79.5 $\pm$ 22.4 $-$   & 82.3 $\pm$ 22.6 $-$  & 125.6 $\pm$ 19.4 $-$ & \textbf{147.0 $\pm$ 27.1} \\
                    & MAZE     & 92.6 $\pm$ 11.6 $-$ & 115.6 $\pm$ 25.7 $-$  & 119.8 $\pm$ 10.9 $-$ & 95.2 $\pm$ 15.7 $-$   & 131.2 $\pm$ 16.4 $-$ & \textbf{164.0 $\pm$ 27.0} \\
                    & Human Proxy  & 28.1 $\pm$ 13.3 $\approx$ & 25.4 $\pm$ 6.8 $\approx$  & 26.9 $\pm$ 7.4 $\approx$   & 28.6 $\pm$ 5.3 $\approx$   & \textbf{35.7 $\pm$ 4.2} $+$ & 25.0 $\pm$ 5.1 \\\midrule
\multicolumn{2}{c|}{$+/-/\approx$}  & 1/14/1   & 1/11/3    & 2/10/4    & 1/8/7   & 2/7/7 & \\
\multicolumn{2}{c|}{Average Rank}  & 5.13   & 4.06    & 4.00    & 3.75    & 2.25 & \textbf{1.81}   \\\bottomrule      
\end{tabular}}
\label{Performance of the test}
\end{table*}

The training curves are shown in Figure~\ref{fig:main fig}, reflecting the change in the average reward of the agents during the training phase. In all the six heterogeneous environments, V-MAZE achieves better performance clearly, showing the necessity of considering the heterogeneity and distinguishing the two players explicitly. Besides, we use a larger scale network to train SP and PP agents on the heterogeneous layout \emph{AA} in Section~\ref{sec:further studies}. With more generations (i.e., almost three times of V-MAZE), SP and PP still perform worse than V-MAZE. This indicates that the heterogeneous skills of both players cannot be mastered well at the same time even with enhanced representation ability and more data, further strengthening our conclusions. \fix{The test performance with unknown partners (defined in Section~\ref{rq2} in Table~\ref{tab:baseline-test} align with the training curves, i.e., V-MAZE $>$ PP $>$ SP}.

\subsection{RQ2: How does MAZE perform compared with recently proposed algorithms?}\label{rq2}

Next, we examine the performance of MAZE in heterogeneous ZSC setting, i.e., testing with unknown partners. Besides SP~\cite{sp,sp2} and PP~\cite{pp}, we compare it with the other three recently proposed algorithms, TrajeDi~\cite{TrajeDi}, FCP~\cite{fcp} and MEP~\cite{MEP}, which use self-play to achieve a diverse partner population. To simulate the performance of coordinating with unseen partners, we test the agents generated by these algorithms with the following partners: 1)~Randomly initialized partner (i.e., Random), which is used to simulate poorly-performed humans; 2)~Self-play partner (i.e., Self-Play), which is trained by SP and used to simulate moderately-performed humans; 3)~Partner trained by MAZE (i.e., MAZE), which is used to simulate well-performed humans. Note that the partner trained by MAZE is accomplished by exchanging the role of agent and partner in the original MAZE and re-training in another separate run (i.e., the MAZE agent and the partner trained by MAZE do not know each other); 4)~Partner trained by behavioral cloning with human data (i.e., Human Proxy), which is used to simulate real humans. The model of Human Proxy is provided in Carroll \textit{et al.}~\cite{overcooked}.

Table~\ref{Performance of the test} shows the detailed results, i.e., the mean and standard deviation of the reward achieved by each algorithm under each combination of layout and partner. We compute the rank of each algorithm under each setting as in~\cite{demsar:06}, which are averaged in the last row of Table~\ref{Performance of the test}. Besides, we apply the Wilcoxon rank-sum test with significance level 0.05, to compare MAZE with other methods. We can observe the order of rank ``MAZE $<$ MEP  $<$ TrajeDi $<$ FCP $<$ PP $<$ SP'', which is consistent with previous observations, e.g., ``FCP $<$ PP $<$ SP'' in Strouse \textit{et al.}~\cite{fcp} and ``MEP  $<$ TrajeDi $<$ FCP $<$ SP" in Zhao \textit{et al.}~\cite{MEP}. As expected, PP using a population is better than SP using a single individual. The superiority of TrajeDi, FCP, and MEP over PP discloses the advantage of exposing the agent to diverse partners during the training process. The proposed method MAZE performs the best overall. Furthermore, its superiority on the four heterogeneous layouts (i.e., \emph{H-CR}, \emph{AA}, \emph{AA-2} and \emph{FC}), suggesting that MAZE is suitable for heterogeneous ZSC. The generally good performance of MAZE with different partners also shows that MAZE can coordinate with partners with different skill levels. We can also observe that for each algorithm on each layout, the partner trained by MAZE is almost always the best. Besides, the MAZE agent paired with the MAZE partner achieves the highest performance on the four heterogeneous layouts. These observations also show the advantage of considering the heterogeneity by MAZE. When paired with Human Proxy partners, the performance of each algorithm is relatively bad, which may be because the human proxy is trained ignoring the heterogeneity and thus hard to coordinate with.

\subsection{RQ3: How do different components of MAZE influence its performance?}\label{rq3}
In this section, we explore the impact of various components of MAZE on its overall performance. Specifically, we examine the effects of pairing strategies, different diversity terms in the updating process, deployment approaches, and conduct comprehensive ablation studies.

\paragraph{Pairing.} As discussed in Section~\ref{sec3.1 pairing}, MAZE employs four pairing strategies, namely fixed, best, worst, and random pairing, to pair an agent with a partner from the partner population. The detailed results of MAZE using these four pairing strategies are presented in Table~\ref{tab-pairing}. Notably, it can be observed that MAZE achieves the best performance when employing random pairing, whereas fixed pairing yields the worst performance. This outcome is in line with expectations, as pairing with a greater variety of partners during training enhances the ability to effectively coordinate with novel and unseen partners. The design of an improved pairing strategy remains an interesting area for further research.

\begin{table*}[t!]
\centering
\caption{The reward (mean$\pm$std.) achieved by MAZE using different pairing strategies. For each combination of layout and partner, the largest reward is bolded.}\label{tab-pairing}
\begin{tabular}{c|c|cccc}
\toprule
     Layout         & Partner     & Fixed    & Best      & Worst   & Random \\ \midrule
\multirow{4}{*}{\emph{H-CR}} & Random      & 4.0 $\pm$ 0.7     & 5.0 $\pm$ 1.4    & 4.5 $\pm$ 0.9   & \textbf{7.0 $\pm$ 0.7}\\
                    & Self-Play   & 50.3 $\pm$ 7.1    & 85.6 $\pm$ 19.9     & \textbf{100.3 $\pm$ 18.3}  & 97.5 $\pm$ 14.6\\
                    & MAZE        & 124.5 $\pm$ 14.2    & 162.7 $\pm$ 20.4  & 174.5 $\pm$ 20.9  & \textbf{183.7 $\pm$ 11.9}\\
                    & Human Proxy & 69.5 $\pm$ 10.1     & 105.0 $\pm$ 19.3     & 98.7 $\pm$ 9.3          & \textbf{118.6 $\pm$ 13.2}\\\midrule
\multirow{4}{*}{\emph{AA}} & Random      & 48.0 $\pm$ 8.7      & 68.9 $\pm$ 15.7      & \textbf{76.0 $\pm$ 16.5}   & 74.4 $\pm$ 15.5    \\
                    & Self-Play   & \textbf{207.0 $\pm$ 20.9} & 162.7 $\pm$ 22.5  & 190.7 $\pm$ 24.5 & 142.6 $\pm$ 20.3               \\
                    & MAZE        & 310.5 $\pm$ 52.6    & 314.7 $\pm$ 18.3     & 306.0 $\pm$ 16.9 & \textbf{315.5 $\pm$ 10.1}            \\
                    & Human Proxy & 43.5 $\pm$ 6.0      & \textbf{127.0 $\pm$ 24.1}   & 123.3 $\pm$ 20.6 & 112.0 $\pm$ 19.1  \\ \midrule
\multirow{4}{*}{\emph{AA-2}}& Random    & 52.0 $\pm$ 10.6     & 53.4 $\pm$ 11.5      & \textbf{58.6 $\pm$ 10.1}   & 56.5 $\pm$ 9.3\\
                    & Self-Play   & 123.4 $\pm$ 29.9    & 110.5 $\pm$ 18.8     & 121.6 $\pm$ 22.2           & \textbf{130.6 $\pm$ 18.5}\\
                    & MAZE        & 308.2 $\pm$ 34.3    & 356.8 $\pm$ 19.7     & 320.5 $\pm$ 16.8  & \textbf{381.4 $\pm$ 13.7}\\
                    & Human Proxy & 60.4 $\pm$ 14.3     & 107.6 $\pm$ 18.4     & 103.4 $\pm$ 16.4  & \textbf{111.5 $\pm$ 
 13.7}\\\midrule
\multirow{4}{*}{\emph{FC}} & Random      & 4.0 $\pm$ 1.9       & 2.5 $\pm$ 1.7       & 4.0 $\pm$ 1.4   & \textbf{7.3 $\pm$ 0.4}\\
                    & Self-Play   & 99.0 $\pm$ 37.9     & 69.5 $\pm$ 19.4     & 106.3 $\pm$ 21.3 & \textbf{147.0 $\pm$ 27.1}      \\
                    & MAZE       & 108.5 $\pm$ 26.0    & \textbf{165.8 $\pm$ 24.1}   & 121.4 $\pm$ 22.8  & 164.0 $\pm$ 27.0    \\
                    & Human Proxy & 18.3 $\pm$ 5.2      & 17.3 $\pm$ 4.8      & 17.3 $\pm$ 3.5  & \textbf{25.0 $\pm$ 5.1}    \\\midrule
                    \multicolumn{2}{c|}{Average Rank} &3.31 & 2.75 & 2.38 & 1.56 \\\bottomrule
\end{tabular}
\end{table*}

\paragraph{Updating.} In the updating process of MAZE, we incorporate a diversity term into the objective function to ensure diversity within both the agent and partner populations, as discussed in Section~\ref{sec3.2 updating}. By default, we utilize Jensen-Shannon Divergence (JSD) to measure the diversity of the policy population. Additionally, we also employ trajectory-based diversity, as introduced in TrajeDi~\cite{TrajeDi}, which calculates diversity based on the trajectory distributions of policies rather than the action distributions used in JSD. 
To examine the impact of the diversity term on experimental results, we introduce MAZE-T, which incorporates trajectory-based diversity as the diversity term of MAZE, and compare it with TrajeDi and the original MAZE. As shown in Table~\ref{tab-trajectory}, we observe that MAZE consistently outperforms the other methods, including MAZE-T and TrajeDi.
Interestingly, despite both TrajeDi and MAZE-T utilizing the same diversity measure, the use of the MAZE framework significantly improves the performance of MAZE-T in comparison to TrajeDi. Moreover, the performance of MAZE and MAZE-T is similar, suggesting that the diversity term has minimal effect on performance in this context. This finding highlights the need for further investigation into the development of diversity measures that are better suited for addressing the ZSC problem in the future.

\begin{table}[t!]
\centering
\caption{The reward (mean$\pm$std.) achieved by TrajeDi, MAZE with trajectory diversity (MAZE-T) and MAZE. For each combination of layout and partner, the largest reward is bolded.}\label{tab-trajectory}
\resizebox{\linewidth}{!}{
\begin{tabular}{c|c|ccc}
\toprule
     Layout         & Partner     & TrajeDi    & MAZE-T      & MAZE  \\ \midrule
\multirow{4}{*}{\emph{H-CR}} & Random      & 5.5 $\pm$ 0.7    & 6.4 $\pm$ 1.3     & \textbf{7.0 $\pm$ 0.7}\\
                    & Self-Play   & 87.6 $\pm$ 9.0    & \textbf{110.2 $\pm$ 14.0}      & 97.5 $\pm$ 14.6\\
                    & MAZE        & 153.8 $\pm$ 16.5    & \textbf{187.5 $\pm$ 14.1}     & 183.7 $\pm$ 11.9\\
                    & Human Proxy & 81.0 $\pm$ 13.3     & 112.7 $\pm$ 16.9        & \textbf{118.6 $\pm$ 13.2}\\\midrule
\multirow{4}{*}{\emph{AA}} & Random      & 49.3 $\pm$ 11.9      & \textbf{78.6 $\pm$ 13.4}      & 74.4 $\pm$ 15.5    \\
                    & Self-Play   & \textbf{180.5 $\pm$ 15.5} & 165.3 $\pm$ 28.4  & 142.6 $\pm$ 20.3               \\
                    & MAZE        & 247.0 $\pm$ 27.1    & 290.4 $\pm$ 27.5     & \textbf{315.5 $\pm$ 10.1}            \\
                    & Human Proxy & 50.2 $\pm$ 6.0      & 100.6 $\pm$ 12.5   & \textbf{112.0 $\pm$ 19.1}  \\ \midrule
\multirow{4}{*}{\emph{AA-2}}& Random    & 45.5 $\pm$ 10.1     & \textbf{58.0 $\pm$ 13.0}       & 56.5 $\pm$ 9.3\\
                    & Self-Play   & 100.8 $\pm$ 12.5    & \textbf{135.8 $\pm$ 17.5}           & 130.6 $\pm$ 18.5\\
                    & MAZE        & 170.5 $\pm$ 11.1    & 356.4 $\pm$ 25.8     & \textbf{381.4 $\pm$ 13.7}\\
                    & Human Proxy & 48.0 $\pm$ 6.9     & 98.7 $\pm$ 16.8      & \textbf{111.5 $\pm$ 13.7} \\\midrule
\multirow{4}{*}{\emph{FC}} & Random      & 6.0 $\pm$ 0.7       & 5.0 $\pm$ 0.9     & \textbf{7.3 $\pm$ 0.4}\\
                    & Self-Play   & 80.8 $\pm$ 22.4     & 144.4 $\pm$ 20.1     & \textbf{147.0 $\pm$ 27.1}      \\
                    & MAZE       & 115.6 $\pm$ 25.7    & 154.7 $\pm$ 25.3   & \textbf{164.0 $\pm$ 27.0}    \\
                    & Human Proxy & 25.4 $\pm$ 6.8      & \textbf{28.0 $\pm$ 8.5}      & 25.0 $\pm$ 5.1   \\\midrule
                    \multicolumn{2}{c|}{Average Rank} &2.75 & 1.69 & 1.56 \\\bottomrule
\end{tabular}
}
\end{table}

\paragraph{Deployment.} We first examine the influence of deployment strategies on the performance of MAZE. When terminating, MAZE gets a population of diverse agents rather than a single one. Table~\ref{tab-deployment} shows the performance of MAZE using two different strategies (i.e., Ensemble and BRO) to deploy the agent population. Given a state, the Ensemble strategy selects the action which receives the most vote from the agents in the population. BRO evaluates each agent in the population using the partners archived by MAZE during training, and selects the best one for testing. We can observe that Ensemble is clearly better than BRO. For each combination of layout and partner, we also test the performance of each agent in the population and report their best and average performance, as shown in the last two columns of Table~\ref{tab-deployment}. The average rank of Ensemble is very close to that of Best (i.e., the true best single agent, which is not known in advance), showing the robustness of the Ensemble strategy.

\begin{table*}[ht!]
\centering
\caption{The reward (mean$\pm$std.) achieved by MAZE using different deployment strategies. For each combination of layout and partner, the largest reward is \textbf{bolded}.}
\begin{tabular}{l|c|cc|cc}
\toprule
     Layout          & Partner   & Ensemble    & BRO   & Best  & Average   \\ \midrule
\multirow{4}{*}{\emph{H-CR}} & Random &7.0 $\pm$ 0.5   & 7.0 $\pm$ 0.3  &\textbf{8.0 $\pm$ 1.0}     & 7.0 $\pm$ 0.5  \\
                    & Self-Play & \textbf{97.5 $\pm$ 12.3}  & 92.4 $\pm$ 11.8    & 96.8 $\pm$ 13.4    & 88.4 $\pm$ 10.2   \\
                    & MAZE      & 183.7 $\pm$ 18.4  & 185.6 $\pm$ 20.3    & \textbf{187.0 $\pm$ 19.0}    & 167.7 $\pm$ 15.3   \\
                    & Human Proxy & \textbf{118.6 $\pm$ 17.5}      &105.0 $\pm$ 10.8   & 117.8 $\pm$ 10.6  & 101.0 $\pm$ 12.5  \\\midrule
\multirow{4}{*}{\emph{AA}} & Random   & \textbf{74.4 $\pm$ 15.5}   & 69.6 $\pm$ 16.8  & 73.0 $\pm$ 18.6     & 68.4 $\pm$ 12.5  \\
                    & Self-Play & 142.6 $\pm$ 20.3     & 147.0 $\pm$ 22.5  & \textbf{148.0 $\pm$ 20.4}    & 111.5 $\pm$ 15.8   \\
                    & MAZE      & 315.5 $\pm$ 10.1  & 300.5 $\pm$ 25.7     & \textbf{318.0 $\pm$ 37.0}    & 243.4 $\pm$ 26.1  \\
                    & Human Proxy& \textbf{112.0 $\pm$ 19.1}  & 108.3 $\pm$ 17.8   & 110.5 $\pm$ 20.0   & 96.1 $\pm$ 12.9    \\\midrule
\multirow{4}{*}{\emph{AA-2}}& Random    & 56.5 $\pm$ 9.3   & 55.0 $\pm$ 12.5   & \textbf{57.8 $\pm$ 11.7}      & 48.2 $\pm$ 7.0 \\
                    & Self-Play & 130.6 $\pm$ 18.5          & 131.8 $\pm$ 20.6   & \textbf{142.0 $\pm$ 22.5}    & 126.2 $\pm$ 19.5 \\
                    & MAZE      & \textbf{381.4 $\pm$ 13.7}  & 342.3 $\pm$ 14.4   & 379.0 $\pm$ 29.8    & 359.9 $\pm$ 18.6  \\
                    & Human Proxy  & \textbf{111.5 $\pm$ 13.7}  & 95.5 $\pm$ 15.2   & 107.0 $\pm$ 16.3 & 99.1 $\pm$ 13.3 \\\midrule
\multirow{4}{*}{\emph{FC}} & Random     & 7.3 $\pm$ 0.4    & 6.5 $\pm$ 1.0   & \textbf{7.5 $\pm$ 1.9}        & 6.0 $\pm$ 0.7  \\
                    & Self-Play & \textbf{147.0 $\pm$ 27.1}  & 126.0 $\pm$ 25.6    & 142.0 $\pm$ 29.6    & 117.0 $\pm$ 22.3 \\
                    & MAZE       & \textbf{164.0 $\pm$ 27.0}  & 149.5 $\pm$ 24.8   & 157.5 $\pm$ 29.0    & 134.4 $\pm$ 21.5\\
                    & Human Proxy& \textbf{25.0 $\pm$ 5.1}    & 22.5 $\pm$ 6.4     & 24.0 $\pm$ 7.4     & 21.0 $\pm$ 6.3\\\midrule
                   \multicolumn{2}{c|}{Average Rank} &1.69  &3.00  & \textbf{1.56}  &3.75  \\\bottomrule
\end{tabular}
\label{tab-deployment}
\end{table*}

\paragraph{Ablation studies.} As we introduce before, MAZE has three main components, i.e., the random pairing of agents and partners (denoted as P), updating each policy by optimizing an objective function with both reward and diversity (denote as D), and partner selection using the archive (denoted as S). To examine the inﬂuence of different components of MAZE, we start from the simplest V-MAZE and add these components gradually, generating a series of variants until the complete MAZE, which are denoted as V-MAZE, V-MAZE$_{+P}$,  V-MAZE$_{+D}$, \fix{V-MAZE$_{+S}$, V-MAZE$_{+S,P}$, V-MAZE$_{+S,D}$,} V-MAZE$_{+P,D}$, and V-MAZE$_{+P,D,S}$ (i.e., MAZE). Note that V-MAZE is the simplest variant of MAZE, which uses the fixed pairing strategy, updates the policies without considering the diversity term in the objective function, and does not use the archive for partner selection. The performance comparison among these methods is shown in Table~\ref{table:ablation of different components}. From the last row, i.e., average rank, we can find that using more components generally yields better performance. V-MAZE$_{+D}$ has a similar performance to V-MAZE, because only adding the diversity term into the objective function without changing pairing cannot ensure that the agents are trained with diverse partners. V-MAZE$_{+P}$ also adds a single component (i.e., random pairing of agents and partners) compared to V-MAZE, but is better than V-MAZE and V-MAZE$_{+D}$. V-MAZE$_{+P}$ changes the partners in pairing, and thus the collected trajectories of each agent-partner pair are more diverse than before, making the agent more likely to coordinate with novel unseen partners well. By combining both the two components, V-MAZE$_{+P,D}$ performs much better. \fix{In V-MAZE$_{+S}$, partners are chosen from the archive each generation, allowing the agent to encounter different partners and improve final coordination ability. V-MAZE$_{+S,P}$ combines selection and pairing, further promoting the agent's exposure to a diverse range of partners. V-MAZE$_{+S,D}$ additionally uses diversity-based updates, making the partners even more diverse and further enhancing the final results.} Finally, the complete MAZE (i.e., V-MAZE$_{+P,D,S}$) has the best average rank, and the largest reward in each heterogeneous environment (i.e., 183.7 in \emph{H-CR}, 315.5 in \emph{AA}, 381.4 in \emph{AA-2}, and 164.0 in \emph{FC}) is all achieved by MAZE. Thus, these results indicate the necessity and effectiveness of each component of the proposed framework.

\begin{table*}[htbp]
\centering
\caption{\fix{The reward (mean$\pm$std.) achieved by MAZE using different components. For each combination of layout and partner, the largest reward is bolded.}}
\resizebox{0.9\linewidth}{!}{
\begin{tabular}{l|c|c|ccc|ccc|c}
\toprule
     Layout         & Partner  & V-MAZE & V-MAZE$_{+P}$ & V-MAZE$_{+D}$ & V-MAZE$_{+S}$ & V-MAZE$_{+S, P}$ & V-MAZE$_{+S, D}$ & V-MAZE$_{+P,D}$ & MAZE \\ \midrule
\multirow{4}{*}{\emph{H-CR}} &Random & 4.0 $\pm$ 0.7  & 4.4 $\pm$ 1.1 & 5.8 $\pm$ 0.8 & 5.4 $\pm$ 0.6 & 5.7 $\pm$ 0.5 & 4.9 $\pm$ 0.6 & 4.5 $\pm$ 0.5  & \textbf{7.0 $\pm$ 0.7} \\
                    &Self Play    & 50.3 $\pm$ 7.1 & 86.1 $\pm$ 6.4& 55.4 $\pm$ 11.2 & 54.5 $\pm$ 10.1 & 58.9 $\pm$ 9.2 & 87.3 $\pm$ 12.4 & \textbf{101.6 $\pm$ 13.1} & 97.5 $\pm$ 14.6\\
                    &MAZE        & 124.5 $\pm$ 14.2 & 152.0 $\pm$ 11.4 & 131.4 $\pm$ 14.6 & 129.7 $\pm$ 12.3 & 145.1 $\pm$ 13.5 & 156.2 $\pm$ 15.1 & 160.2 $\pm$ 14.5  & \textbf{183.7 $\pm$ 11.9} \\
                    &Human Proxy & 69.5  $\pm$ 10.1 & 94.6 $\pm$ 18.9 & 72.0 $\pm$ 13.0  & 86.1 $\pm$ 11.5 & 76.2 $\pm$ 12.4 & 103.4 $\pm$ 14.8 & 101.8 $\pm$ 16.3 & \textbf{118.6 $\pm$ 13.2} \\\midrule
\multirow{4}{*}{\emph{AA}} &Random & 48.0 $\pm$ 8.7  & 52.0 $\pm$ 13.5  & 57.0 $\pm$ 16.1   & 55.4 $\pm$ 9.3 & 57.3 $\pm$ 12.1 & 61.2 $\pm$ 11.7 & 60.3 $\pm$ 10.9  & \textbf{74.4 $\pm$ 15.5}  \\
                    & Self Play   & \textbf{207.0 $\pm$ 20.9} & 151.7 $\pm$ 36.9 & 122.0 $\pm$ 23.4 & 178.1 $\pm$ 33.4  & 145.1 $\pm$ 31.2 & 182.3 $\pm$ 27.9 & 168.5 $\pm$ 35.1  & 142.6 $\pm$ 20.3 \\
                    &MAZE     & 310.5 $\pm$ 52.6 & 286.0 $\pm$ 16.1  & 262.2 $\pm$ 37.4 & 251.7 $\pm$ 32.1 & 249.8 $\pm$ 30.0 & 301.2 $\pm$ 21.4 & 311.5 $\pm$ 17.4  & \textbf{315.5 $\pm$ 10.1} \\
                    &Human Proxy  & 43.5 $\pm$ 6.0  & 101.3 $\pm$ 17.5  & 82.5 $\pm$ 22.2  & 113.5 $\pm$ 21.7 & 117.2 $\pm$ 17.2 & 122.3 $\pm$ 16.7 & \textbf{126.3 $\pm$ 27.3}  & 112.0 $\pm$ 19.1 \\\midrule
\multirow{4}{*}{\emph{AA-2}}&Random & 52.0 $\pm$ 10.6  & 59.6 $\pm$ 20.2  & 49.8 $\pm$ 12.8  & 49.1 $\pm$ 11.3 &  \textbf{62.3 $\pm$ 21.1} & 59.3 $\pm$ 19.4 & 53.0 $\pm$ 11.4   & 56.5 $\pm$ 9.3\\
                    & Self Play    & 123.4 $\pm$ 29.9& 100.8 $\pm$ 25.4 & 117.0 $\pm$ 27.8 & 106.3 $\pm$ 18.9 & 122.5 $\pm$ 26.7 & 129.3 $\pm$ 17.7 & 127.5 $\pm$ 27.3  & \textbf{130.6 $\pm$ 18.5} \\
                    &MAZE         & 308.2 $\pm$ 34.3    & 362.5 $\pm$ 29.2 & 327.3 $\pm$ 36.4 & 328.2 $\pm$ 27.6 & 345.7 $\pm$ 26.8 & 352.3 $\pm$ 31.7 & 343.0 $\pm$ 29.5  & \textbf{381.4 $\pm$ 13.7} \\
                    &Human Proxy  & 60.4 $\pm$ 14.3  & 75.3 $\pm$ 19.8  & 80.8 $\pm$ 18.0   & 81.4 $\pm$ 19.4  & 87.7 $\pm$ 15.1 & 103.7 $\pm$ 16.3 & 104.6 $\pm$ 12.5  & \textbf{111.5 $\pm$ 13.7} \\\midrule
\multirow{4}{*}{\emph{FC}} &Random  & 4.0 $\pm$ 1.9  & 2.7 $\pm$ 1.1   & 3.0 $\pm$ 2.0     & 2.3 $\pm$ 1.2 & 3.6 $\pm$ 1.7 & 4.5 $\pm$ 0.6 & 4.0 $\pm$ 0.4    & \textbf{7.3 $\pm$ 0.4}    \\
                    &Self Play    & 99.0 $\pm$ 37.9  & 123.0 $\pm$ 34.1 & 79.4 $\pm$ 24.0 & 82.1 $\pm$ 34.7 & 121.3 $\pm$ 27.2 & 132.1 $\pm$ 29.4 & 113.3 $\pm$ 35.5  & \textbf{147.0 $\pm$ 27.1} \\
                    &MAZE       & 108.5 $\pm$ 26.0 & 131.8 $\pm$ 21.4 & 93.4 $\pm$ 20.0  & 112.5 $\pm$ 18.7 & 114.3 $\pm$ 17.9 & 119.1 $\pm$ 28.2 & 123.0 $\pm$ 29.1  & \textbf{164.0 $\pm$ 27.0} \\
                    &Human Proxy  & 18.3 $\pm$ 5.2  & 11.3 $\pm$ 4.7 & 17.8 $\pm$ 2.6   & 15.4 $\pm$ 3.2 & 20.3 $\pm$ 4.9 & 17.9 $\pm$ 5.1 & 16.3 $\pm$ 4.7   & \textbf{25.0 $\pm$ 5.1} \\
                    \midrule
                   \multicolumn{2}{c|}{Average Rank} & 6.13 & 5.06 & 6.25  & 6.06 & 4.37 & 2.88 & 3.37 & \textbf{1.87} \\ \bottomrule
\end{tabular}}
\label{table:ablation of different components}
\end{table*}

\fix{\paragraph{Sensitivity analysis.} MAZE has two hyper-parameters: population size $n_P$ and and archive size $n_A$, which is set to $5$ and $20$ by default. We further conduct experiments to analysis their influence on the final performance.
We compare the default MAZE with MAZE-pop3 (i.e., the population size is set to 3, with the archive size remaining unchanged at 20), MAZE-pop7, MAZE-archive10 (i.e., the archive size is set to 10, with the population size remaining unchanged at 5), and MAZE-archive40. The results can be found in Table~\ref{tab:sensitivity-analysis}. 
For a fair experimental comparison, we keep the number of updates for each agent consistent. Therefore, changing the population size will result in different total number of updates, which may allow larger populations to achieve relatively better results (i.e., the more agents we have, the easier it is to find better-performing agents). When the archive size increases, the number of updates for each potential partner decreases, leading to slightly lower performance in the same number of iterations. However, compared to other comparison algorithms in the main experiments (i.e., Table~\ref{Performance of the test}), all the MAZE variants still maintain good performance.}

\begin{table*}[t!]
\centering
\caption{\fix{The reward (mean$\pm$std.) achieved by MAZE with different population size and archive size.  MAZE uses population size 5 and archive size 20 by default. MAZE-pop3 and MAZE-pop7 denote MAZE with population size 3 and 7, respectively; MAZE-archive10 and MAZE-archive40 denote MAZE with archive size 10 and 40, respectively. For each combination of layout and partner, the largest reward is bolded.}}
\resizebox{0.9\linewidth}{!}{
\begin{tabular}{c|c|c|cc|cc}
\toprule
     Layout         & Partner     & MAZE      & MAZE-pop3 & MAZE-pop7 & MAZE-archive10 & MAZE-archive40 \\ \midrule
\multirow{4}{*}{\emph{AA-2}}& Random    & 56.5 $\pm$ 9.3 & \textbf{59.3 $\pm$ 8.5} & 49.2 $\pm$ 7.3 & 55.2 $\pm$ 7.1 & 53.4 $\pm$ 9.1 \\
                    & Self-Play   & \textbf{130.6 $\pm$ 18.5} & 127.2 $\pm$ 14.7 & 121.2 $\pm$ 17.9 & 124.5 $\pm$ 14.7 & 119.3 $\pm$ 14.6 \\
                    & MAZE        & 381.4 $\pm$ 13.7 & 382.8 $\pm$ 11.2 & \textbf{389.1  $\pm$ 14.1} & 371.0  $\pm$ 12.5  & 376.4 $\pm$ 8.6 \\
                    & Human Proxy & 111.5 $\pm$ 13.7 & 109.4 $\pm$ 16.8 & 98.4 $\pm$ 7.3 & \textbf{113.2 $\pm$ 12.9} & 87.2 $\pm$ 9.6 \\\midrule
\multirow{4}{*}{\emph{FC}} & Random      & 7.3 $\pm$ 0.4 & 5.2 $\pm$ 0.1 & \textbf{7.7 $\pm$ 0.2} & 7.4 $\pm$ 0.7 & 6.1 $\pm$ 0.9 \\
                    & Self-Play   & \textbf{147.0 $\pm$ 27.1} & 138.1 $\pm$ 26.3 & 146.1 $\pm$ 20.4 & 141.2 $\pm$ 19.8 & 132.0 $\pm$ 21.5\\
                    & MAZE       & \textbf{164.0 $\pm$ 27.0} & 153.2 $\pm$ 25.4 & 161.2 $\pm$ 26.1 & 165.1  $\pm$ 27.7 & 156.9  $\pm$  22.1 \\
                    & Human Proxy & 25.0 $\pm$ 5.1 & 19.2 $\pm$ 4.8 & \textbf{27.1 $\pm$ 6.3} & 26.1 $\pm$ 4.2 & 20.4 $\pm$ 6.3 \\\midrule
\multicolumn{2}{c|}{Average Rank} & \textbf{2.13} & 3.37 & 2.63 & 2.5 & 4.37 \\\bottomrule
\end{tabular}
}\label{tab:sensitivity-analysis}
\end{table*}

\subsection{RQ4: Can MAZE coordinate well with real human partners?}\label{sec-rq4}
In order to investigate the coordination abilities of the algorithms with real humans, we conduct experiments involving human participants. A total of 8 participants are recruited for the evaluation of three algorithms, i.e., FCP, MEP, and MAZE, on the \emph{AA} layout of \emph{Overcooked}. To mitigate any potential experimental biases, each participant experienced all three algorithms in a randomized order.

To address the potential risks and ethical concerns associated with the real human experiments, we took several measures. Prior to participating in the experiments, all human participants were provided with detailed information about the nature of the experiments. This included an explanation of their role in playing the game with AI agents and the intended use of their game-play data for academic research purposes. It is important to note that all collected data was strictly used for experimental purposes only. Furthermore, participation in the experiments was voluntary, and we ensured that the participants were fully informed of our practices and provided their agreement to participate.

\begin{figure}[htbp]
    \centering
    \includegraphics[width=0.45 \textwidth]{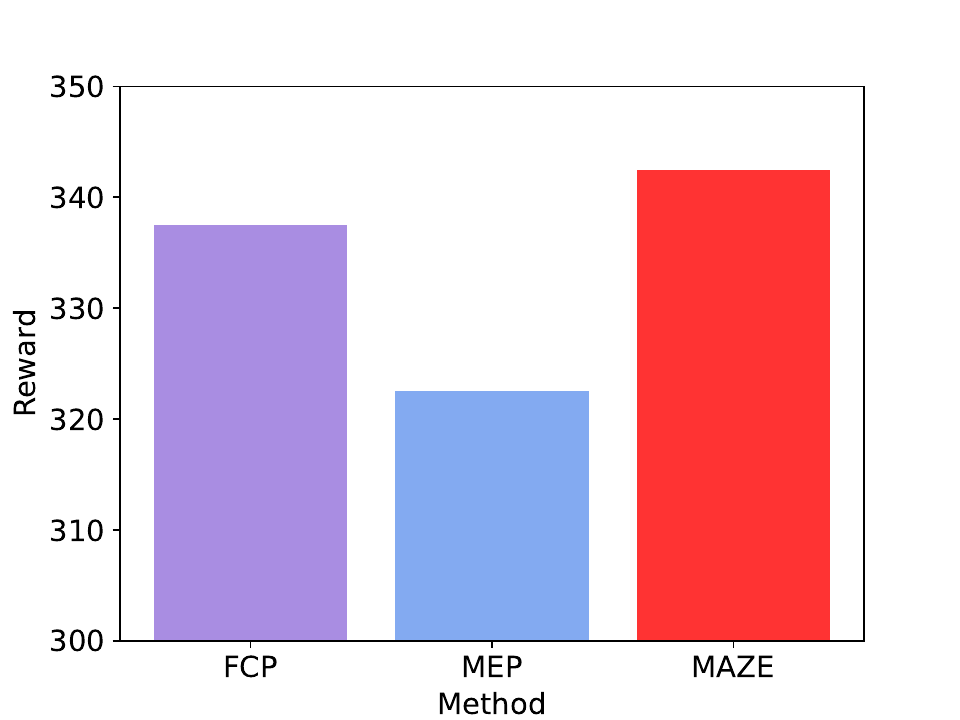}
    \caption{The reward achieved by FCP, MEP, and MAZE when testing with real human participants on the heterogeneous \emph{AA} layout from the \emph{Overcooked} environment. }
    \label{fig:real-human}
\end{figure}

The results of the environment reward achieved when coordinating with real human participants are depicted in Figure~\ref{fig:real-human}. It is worth noting that, due to the superior understanding of the environment and the task exhibited by real humans, the rewards obtained when coordinating with them using all three algorithms surpass those achieved when coordinating with human proxies. Among the three baseline methods, MAZE is the best one, which aligns with the findings from RQ2.

\subsection{Further Studies}\label{sec:further studies}
We include some further studies in this section, including the influence of large network, visualization of partner diversity, running time analysis of MAZE, and the comparison between homogeneous and heterogeneous environments.

\begin{figure}[htbp]
    \centering
    \includegraphics[width=0.4 \textwidth]{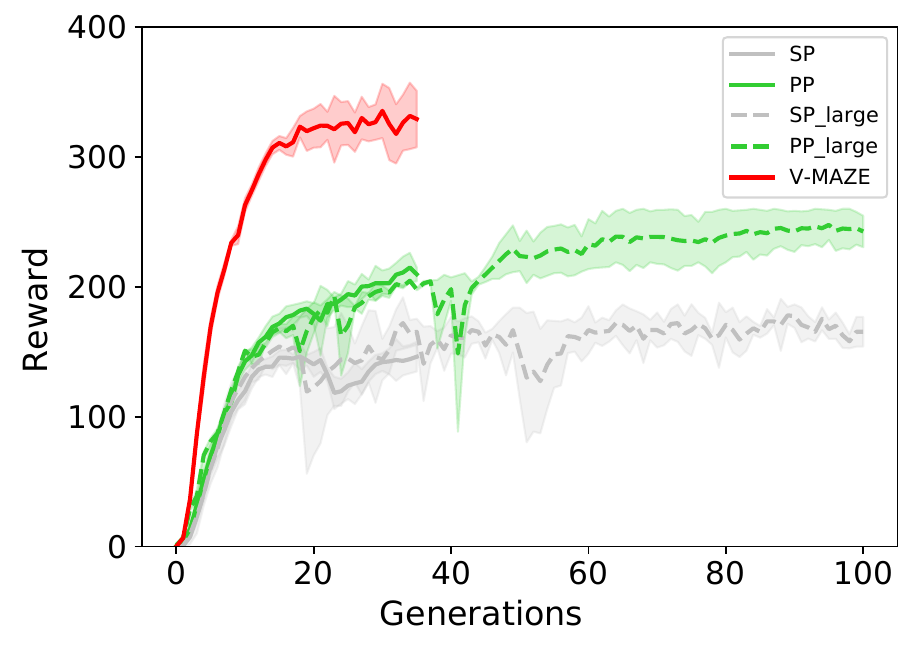}
    \caption{Training curves of SP, PP, SP\_large, PP\_large, and V-MAZE on the heterogeneous \emph{AA} layout from the \emph{Overcooked} environment. }
    \label{fig:larger_network}
\end{figure}

\paragraph{Influence of Larger Network.} In general, larger networks have better representation abilities. We want to answer the following question: \emph{Can a single larger scale network with more data capture the heterogeneous skills?} We use a larger scale network (i.e., the size of hidden layers from 64 to 128) with more number of generations (i.e., from 35 to 100) to train SP and PP agents on the heterogeneous \emph{AA}, obtaining SP\_large and PP\_large. As shown in Figure~\ref{fig:larger_network}, the performance improvement of SP\_large and PP\_large is very limited, indicating that even with enhanced representation ability and more data, the heterogeneous skills of both players cannot be mastered well at the same time. This result verifies the effectiveness and efficiency of MAZE, demonstrating that MAZE is a promising solution for heterogeneous cooperative MARL.

\begin{figure}[htbp]
    \centering
    \includegraphics[width=0.4\textwidth]{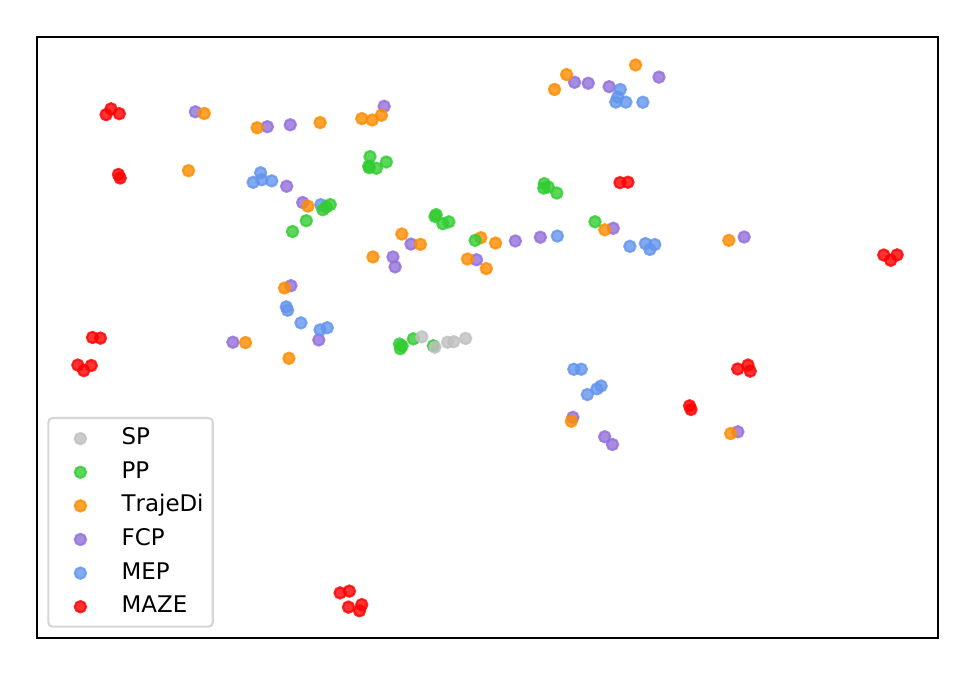}
    \caption{Visualization of partner diversity. We show the action trajectory of the partners from different algorithms by t-SNE embedding in the same space.}
    \label{fig:diversity}
\end{figure}

\paragraph{Visualization of Partner Diversity.} Recent work shows that exposing the agent to diverse partners during the training process efficiently improves the ability to coordinate with unseen partners~\cite{TrajeDi,fcp,MEP}. To verify if the obtained partners are diverse, we visualize the population of partners obtained by different methods after the training phase. For a fair comparison, we pair each partner with the same agent and record the partner's action trajectory. We repeat this five times and 
then visualize all the obtained trajectories in a two-dimensional space using t-SNE~\cite{tsne}. The result is shown in Figure~\ref{fig:diversity}. Note that there are 25 trajectories (i.e., 25 points) for the population-based methods (five partners obtained by each method and five trajectories for each partner). \fix{Some points are clustered nearby in the figure, because we repeated the experiments with same agents multiple times to obtain multiple training trajectories.} The partners obtained by MAZE are the most diverse because they are distributed in different areas of the embedded space. xxThe partners obtained by TrajeDi, FCP, and MEP are also distributed to some extent but not as diverse as MAZE. This is because their mechanisms for maintaining partner diversity are less effective than MAZE. As a population-based method, PP's partners are not diverse enough due to the lack of explicit mechanisms to maintain diversity. The partners obtained by SP are concentrated, which is consistent with its worst performance in RQ1.

\paragraph{Running time analysis of MAZE.} In addition, we compute the wall-clock time for different processes of MAZE on the \emph{AA} layout, specifically the collecting trajectories, updating, and pairing, resulting in 150 seconds, 82 seconds, and approximately 0 seconds, respectively. The primary factor contributing to the overall time overhead is the process of collecting trajectories and updating, which are shared processes among the ZSC methods. Consequently, the time overhead associated with MAZE is comparable to that of other ZSC methods.

\begin{figure}[htbp]
    \centering
    \includegraphics[width=0.47\linewidth]{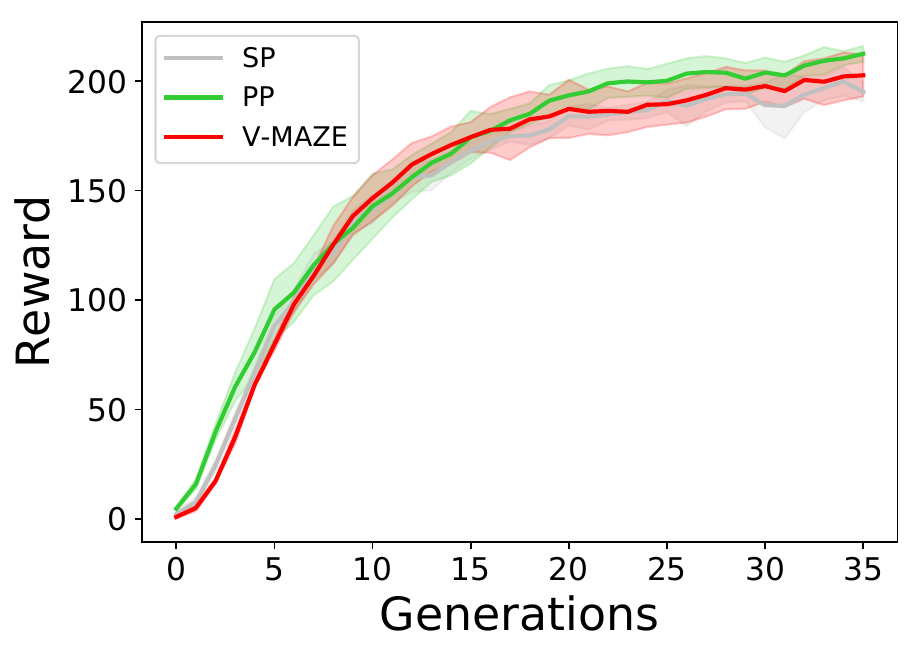}
    \includegraphics[width=0.47\linewidth]{fig/simple-2.pdf} 
    \begin{minipage}[c]{0.47\linewidth}\centering
    \small (a) \emph{CR}
    \end{minipage}
    \begin{minipage}[c]{0.47\linewidth}\centering
    \small (d) \emph{H-CR}
    \end{minipage}
    \caption{Training curves on \emph{CR} and \emph{H-CR}.}\label{cr and h-cr}
\end{figure}

\paragraph{Comparison between homogeneous and heterogeneous environments.}
Finally, we compare the performance of different methods on a homogeneous environment \emph{CR} and a heterogeneous environment \emph{H-CR}. The layouts of these environments are the same. However, the skills of different players in \emph{CR} are the same, and the skills in \emph{H-CR} are different.  Similar to Section~\ref{rq1}, we first show the training curves of V-MAZE, SP, and PP, as shown in Figure~\ref{cr and h-cr}. As expected, on the homogeneous layout \emph{CR}, V-MAZE achieves similar performance to SP and PP. While on the heterogeneous layout \emph{H-CR}, V-MAZE achieves better performance clearly, showing the superior performance of V-MAZE on the heterogeneous environment.

\section{Conclusion}
This paper considers the problem of ZSC with unseen partners in MARL. Previous works mainly focused on the homogeneous setting, while the heterogeneous one also widely exists in practice. We first highlight the importance of explicitly considering the heterogeneity of multi-agent coordination, then propose a novel framework MAZE based on coevolution, consisting of the three sub-processes of pairing, updating and selection. Comprehensive experiments on two environments with various degrees of heterogeneity demonstrate the superior performance of MAZE. 

MAZE lays the foundation to train the heterogeneous representation of agent and partner, where the population-based self-play methods can not be used. Heterogeneous representation is more suitable for heterogeneous ZSC tasks, which we will consider as our future work. The simple random pairing strategy works well in MAZE, which may be further improved by multi-objective optimization~\cite{hao2024enhancing,liang2024evolutionary,mo-selection} and curriculum learning~\cite{curriculum-short-survey}. Besides, we will try to test the performance of MAZE on more types of environments (e.g., Hanabi~\cite{ZSC}) and analyze the advantage of MAZE from the theoretical perspective~\cite{heterogeneous-jmlr,hu2023heterogeneous,qd-theory}.

\bibliographystyle{ieee}
\bibliography{ieee_main}

\begin{IEEEbiography}[{\includegraphics[width=1in,height=1.25in,clip,keepaspectratio]{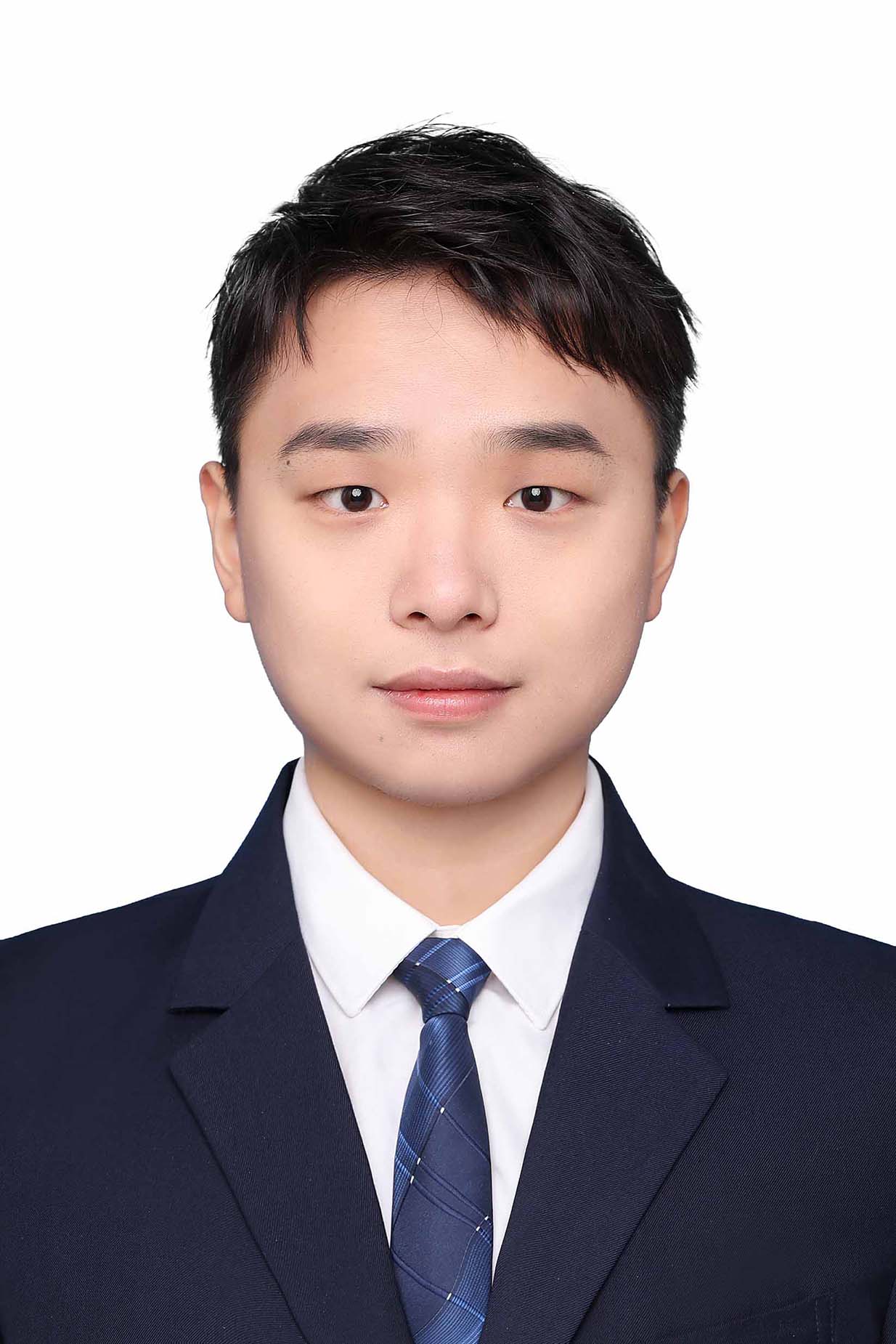}}]{Ke Xue}
received the BSc degree in Mathematics and Applied Mathematics from School of Mathematics, Sun Yat-Sen University in 2019. He is currently pursuing the PhD degree with the School of Artificial Intelligence, Nanjing University, Nanjing, China. His research interests mainly include machine learning and evolutionary computation.
\end{IEEEbiography}

\begin{IEEEbiography}[{\includegraphics[width=1in,height=1.25in,clip,keepaspectratio]{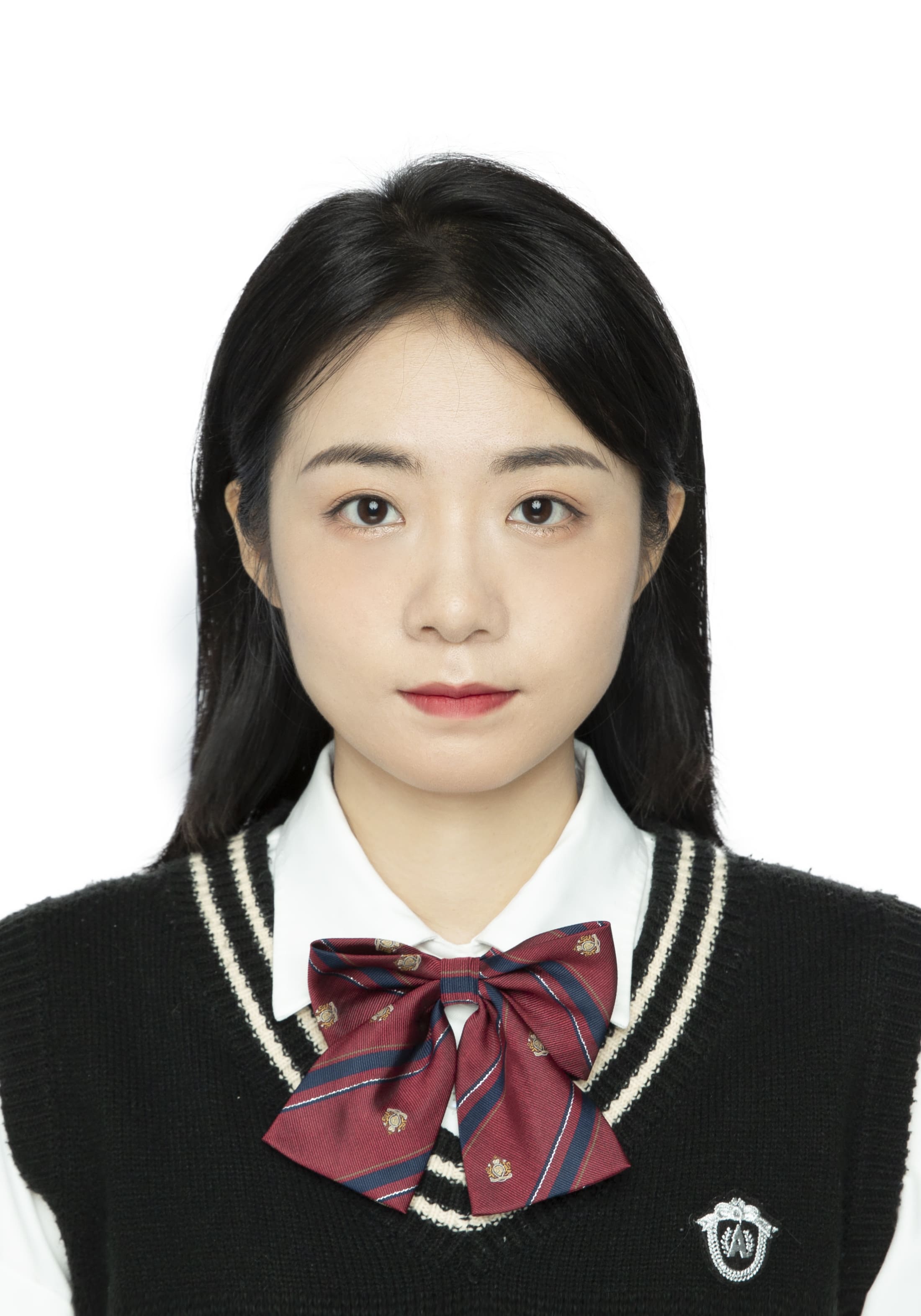}}]{Yutong Wang}
received her BSc degree from the School of Mathematical Sciences, University of Science and Technology of China in 2020 and her MSc degree from the School of Artificial Intelligence, Nanjing University in 2023. She is currently working as an engineer at Huawei Noah’s Ark Lab, Shenzhen, China. Her research mainly interests include machine learning, reinforcement learning and evolutionary algorithms.
\end{IEEEbiography}

\begin{IEEEbiography}[{\includegraphics[width=1in,height=1.25in,clip,keepaspectratio]{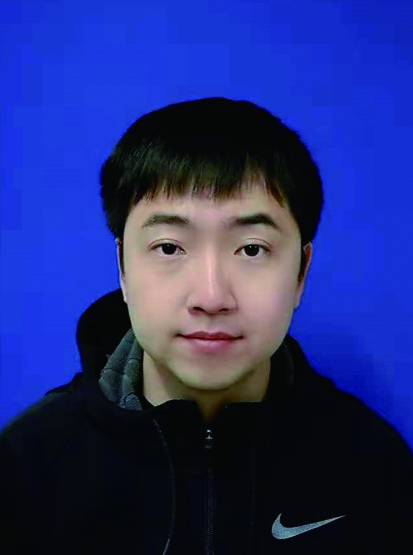}}]{Cong Guan}
received the BSe degree and MSc degree from State Key Laboratory of Synthetical Automation, Northeastern University, and PhD degree from School of Artificial Intelligence, Nanjing University. His research interests mainly include machine learning, reinforcement learning, and multi-agent reinforcement learning.
\end{IEEEbiography}

\begin{IEEEbiography}[{\includegraphics[width=1in,height=1.25in,clip,keepaspectratio]{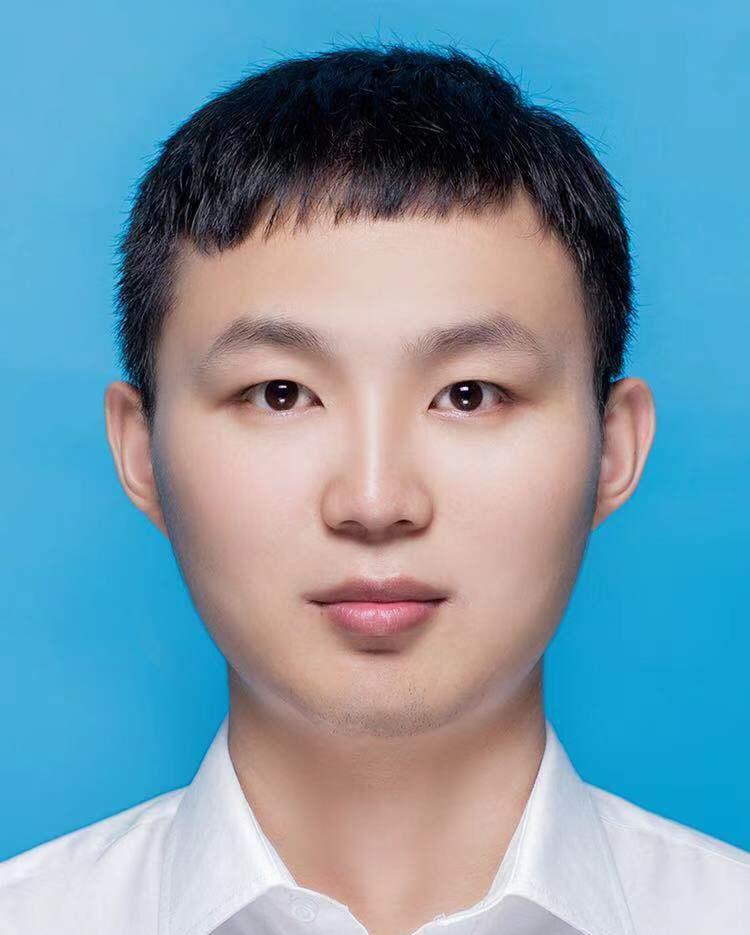}}]{Lei Yuan}
received the BSc degree in Department of Electronic Engineering from Tsinghua University, MSc degree from Chinese Aeronautical Establishment, and PhD degree from Nanjing University.
He is currently an assistant researcher in School of Artificial Intelligence at Nanjing University.
His research interests mainly include machine learning, reinforcement learning, and multi-agent reinforcement learning.
\end{IEEEbiography}

\begin{IEEEbiography}[{\includegraphics[width=1in,height=1.25in,clip,keepaspectratio]{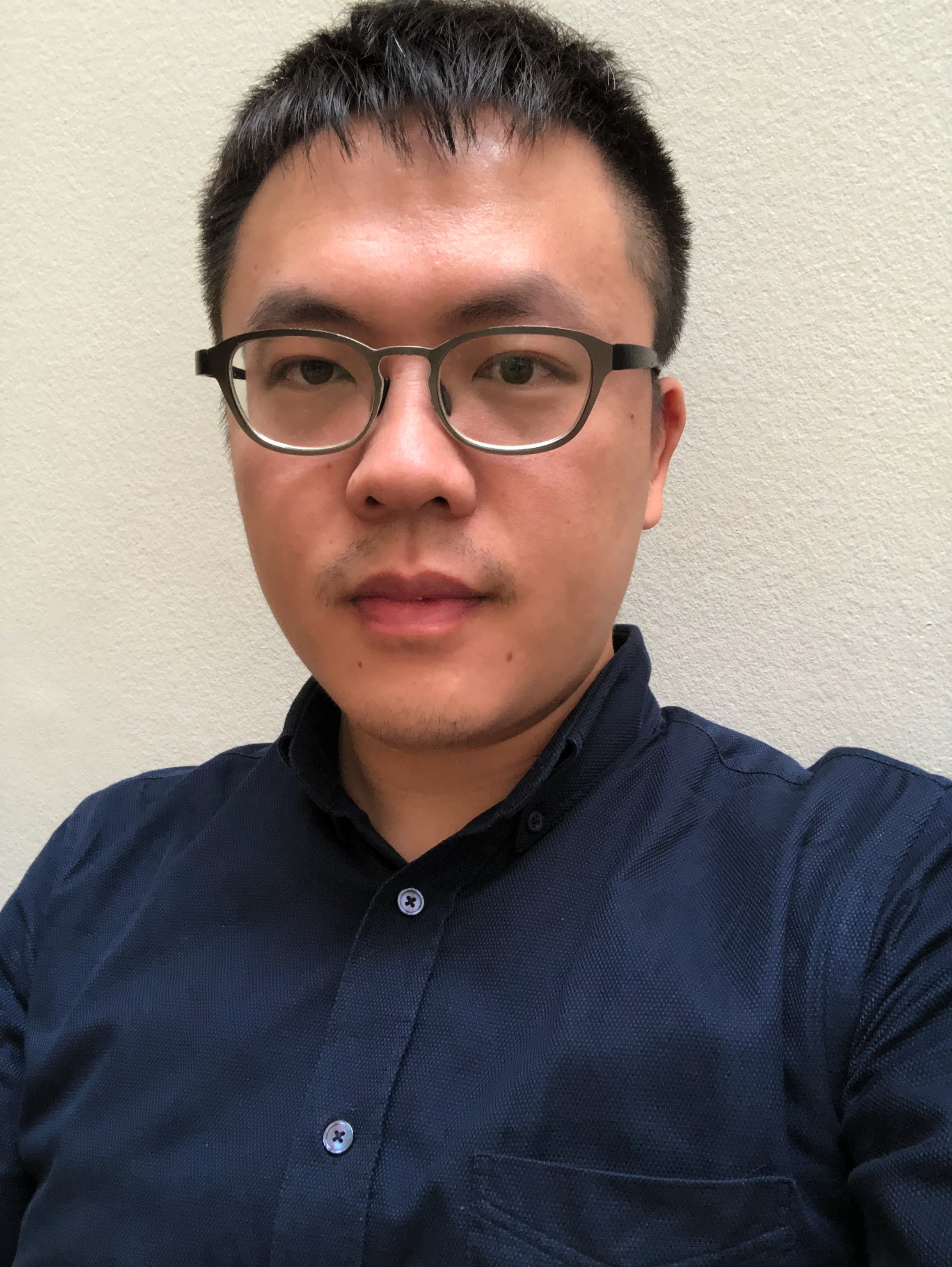}}]{Haobo Fu} received the PhD degree in computer science from University of Birmingham, Birmingham, U.K., in 2014. From 2014 to 2016, he was a senior researcher with the Big Data
Laboratory, Baidu, Inc., Beijing, China. Since 2016, he has been working in the AI Platform Department, Tencent Inc., Shenzhen, China. He is now a tech lead and a principal researcher. His main interests are new innovations in AI for the game industry. These include innovative game AI applications and related research topics, such as machine learning, reinforcement learning, game theory, evolutionary computation, etc.  He has authored or co-authored over 40 refereed publications.
\end{IEEEbiography}

\begin{IEEEbiography}[{\includegraphics[width=1in,height=1.25in,clip,keepaspectratio]{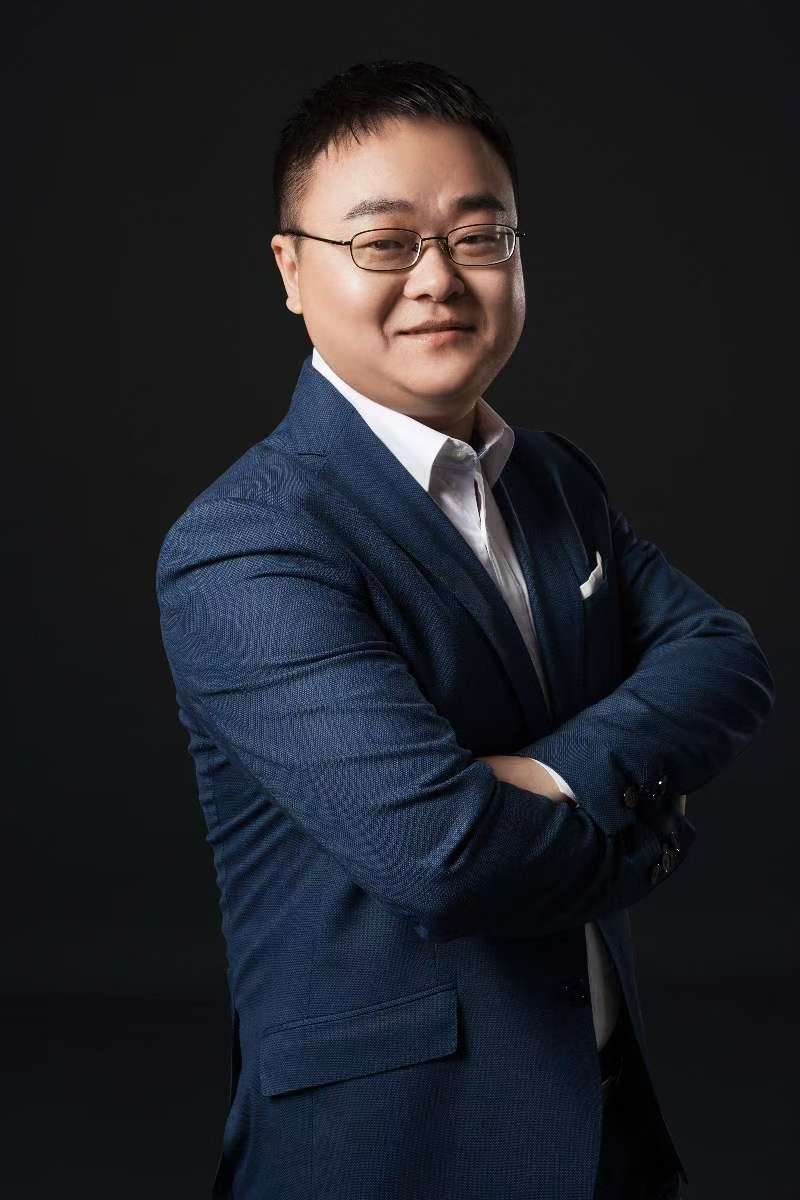}}]{Qiang Fu} is currently the Director of Game AI center, Tencent AI Lab. His research interest includes reinforcement learning, domain data mining and multiagent systems. Leon's current focus is leading the game AI R\&D team to study intelligent AI in games and corresponding applications using deep learning, reinforcement learning and game theory.
\end{IEEEbiography}
\begin{IEEEbiography}[{\includegraphics[width=1in,height=1.25in,clip,keepaspectratio]{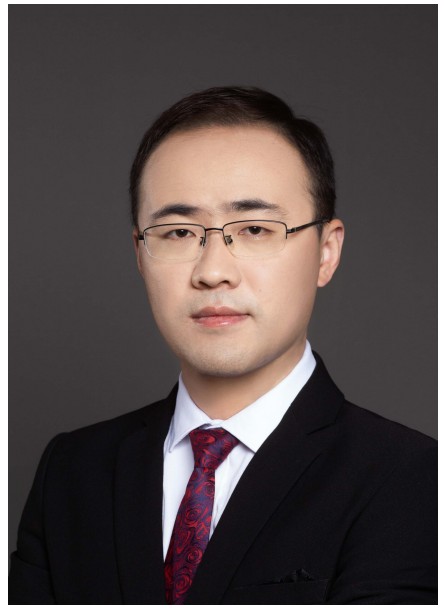}}]{Chao Qian} 
(Senior Member, IEEE) is an Professor in the School of Artificial Intelligence, Nanjing University, China. He received the BSc and PhD degrees in the Department of Computer Science and Technology from Nanjing University in 2009 and 2015, respectively. His research interests include artificial intelligence, evolutionary computation, and machine learning. He has published one book “Evolutionary Learning: Advances in Theories and Algorithms”, and over 50 first/corresponding-authored papers in top-tier journals (PNAS, AIJ, ECJ, TEVC, Algorithmica, TCS) and conferences (AAAI, IJCAI, ICML, NeurIPS, ICLR). He has won the ACM GECCO 2011 Best Theory Paper Award, IDEAL 2016 Best Paper Award, IEEE CEC 2021 Best Student Paper Award Nomination, and Huawei Spark Award twice. He is an editorial board member of Artificial Intelligence Journal and Evolutionary Computation Journal, an associate editor of IEEE Transactions on Evolutionary Computation and IEEE Computational Intelligence Magazine, a young associate editor of Science China Information Sciences, the founding chair of IEEE Computational Intelligence Society (CIS) Task Force on Evolutionary Learning, and was also the chair of IEEE CIS Task Force on Theoretical Foundations of Bio-inspired Computation. He has been invited to give an Early Career Spotlight Talk at IJCAI 2022, and will be a Program Co-Chair of PRICAI 2025. He is a recipient of the National Science Foundation for Excellent Young Scholars (2020), and CCF-IEEE CS Young Computer Scientist Award (2023), and has hosted a National Science and Technology Major Project.
\end{IEEEbiography}

\begin{IEEEbiography}[{\includegraphics[width=1in,height=1.25in,clip,keepaspectratio]{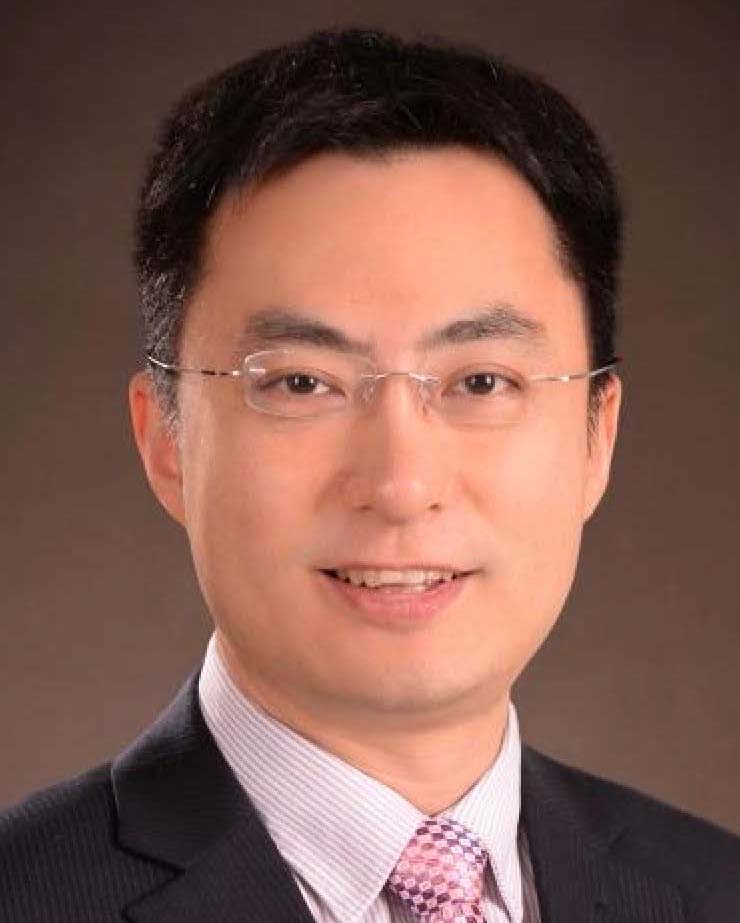}}]{Yang Yu}
(Senior Member, IEEE) received the  PhD degree in the Department of Computer Science and Technology from Nanjing University in 2011, and is currently a Professor at the School of Artificial Intelligence, Nanjing University. His research interests include machine learning, mainly reinforcement learning and derivative-free optimization for learning. Prof. Yu was granted the CCF-IEEE CS Young Scientist Award in 2020, recognized as one of the AI’s 10 to Watch by IEEE Intelligent Systems, and received the PAKDD Early Career Award in 2018.  His teams won the Champion of the 2018 OpenAI Retro Contest on transfer reinforcement learning and the 2021 ICAPS Learning to Run a Power Network Challenge with Trust. He served as Area Chairs for NeurIPS, ICML, IJCAI, AAAI, etc.
\end{IEEEbiography}
\end{document}